\documentclass[10pt,twocolumn,letterpaper]{article}

\usepackage[pagenumbers]{cvpr} 

\usepackage[dvipsnames]{xcolor}

\usepackage{lipsum}
\definecolor{cvprblue}{rgb}{0.21,0.49,0.74}
\usepackage[pagebackref,breaklinks,colorlinks,citecolor=cvprblue]{hyperref}
\usepackage{bbm}

\def\paperID{6632} 
\def\confName{CVPR}
\def\confYear{2024}

\usepackage{inconsolata}
\usepackage[textsize=footnotesize]{todonotes}
\usepackage{booktabs}
\usepackage{multirow}
\usepackage{enumitem}

\usepackage{pifont}
\newcommand{\aspace}{\hspace{1em}}
\newcommand{\usc}{$^{1}$}
\newcommand{\uw}{$^{2}$}
\newcommand{\biu}{$^{3}$}
\newcommand{\google}{$^{4}$}
\usepackage{tikz}
\newcommand{\dreamicon}{\includegraphics[scale=0.02]{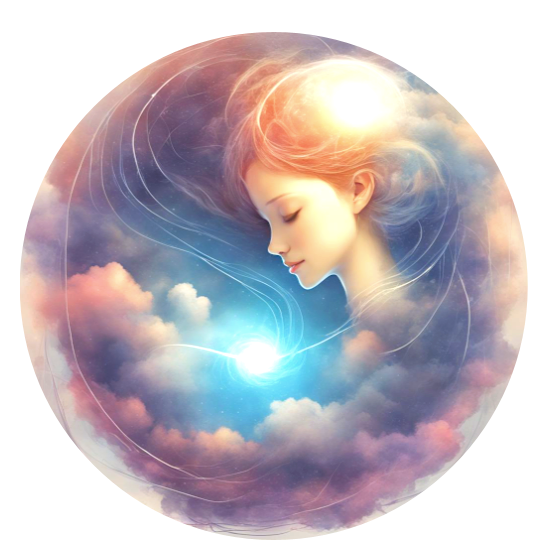}}
\newcommand{\ours}{DreamSync\xspace}

\usepackage{etoolbox}
\makeatletter
\let\@titlehook=\relax
\apptocmd{\@maketitle}{\@titlehook}{}{}
\newcommand{\titlehook}[1]{\def\@titlehook{#1}}
\makeatother

\definecolor{gred}{RGB}{234,67,53}
\definecolor{ggreen}{RGB}{52,168,83}
\newcommand{\score}[2]{{#1}}

\newcommand{\scoreg}[2]{{#1} {\color{ggreen}{({#2})}}}
\newcommand{\scorel}[2]{{#1} {\color{gred}{({#2})}}}
\usepackage{multicol}
\usepackage{multirow}
\usepackage{xcolor,colortbl}
\usepackage{hhline}
\definecolor{Gray}{gray}{0.9}
\newcolumntype{g}{>{\columncolor{Gray}}c}

\usepackage{float}
\usepackage{tikz}
\usepackage{array}
\newcolumntype{P}[1]{>{\centering\arraybackslash}p{#1}}
\newcolumntype{M}[1]{>{\centering\arraybackslash}m{#1}}
\newcolumntype{C}[1]{>{\centering\let\newline\\\arraybackslash\hspace{0pt}}m{#1}}
\usepackage{makecell}

\renewcommand{\paragraph}[1]{
  \vspace{0.2ex} \noindent\textbf{#1}
}

\newcommand{\best}[1]{{\textbf{#1}}}

\usepackage{tablefootnote}

\newcommand{\bfit}[1]{\textbf{\textit{#1}}}

\title{\ours:  Aligning Text-to-Image Generation  with \\ Image Understanding Feedback}
\author{
\textbf{Jiao Sun}$^{1*}$ \aspace 
\textbf{Deqing Fu}$^{1*}$ \aspace 
\textbf{Yushi Hu}$^{2}$\thanks{Equal Contribution. Work done while at Google.} \aspace 
\textbf{Su Wang}\google \aspace 
\textbf{Royi Rassin}\biu \\
\textbf{Da-Cheng Juan}\google \aspace
\textbf{Dana Alon}\google\aspace
\textbf{Charles Herrmann}\google \aspace 
\textbf{Sjoerd van Steenkiste}\google \\
\textbf{Ranjay Krishna}\uw \aspace  
\textbf{Cyrus Rashtchian}\google \\ 
\usc University of Southern California \aspace \uw University of Washington \aspace \biu Bar-Ilan University \\ \google Google Research
}

\begin{document}

\titlehook{
\vspace{-32pt}
\begin{center}
    \centering
    \captionsetup{type=figure}
    \includegraphics[width=0.85\linewidth]{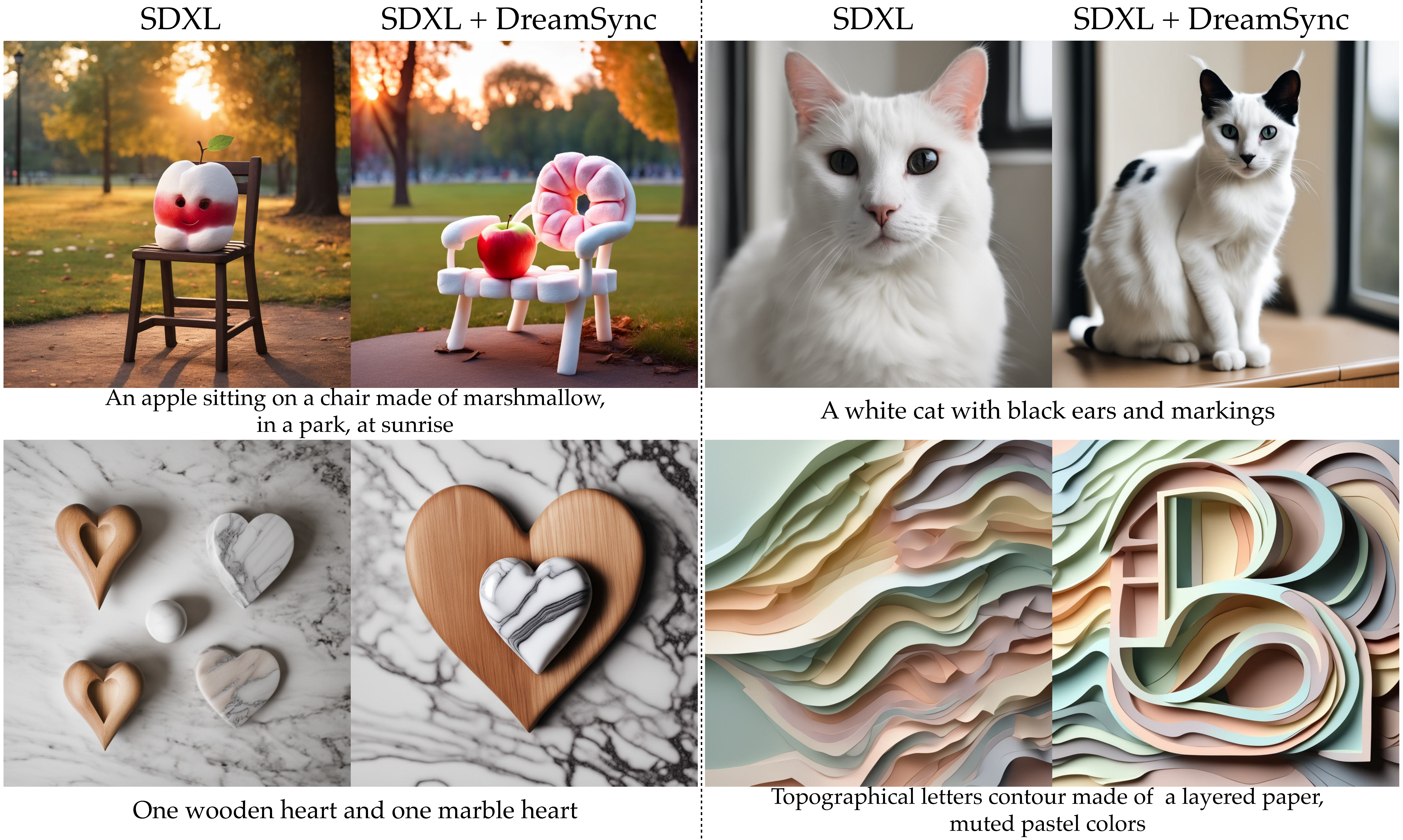}
    \vspace{-8pt}
    \captionof{figure}{We introduce \textbf{\ours}: a model-agnostic training algorithm that improves text-to-image (T2I) generation models' faithfulness to text inputs and image aesthetics. 
    \ours learns from feedback of vision-language models (VLMs), and does not need any human annotation, model architecture changes, or reinforcement learning.
    }
    
\end{center}%
}

\maketitle

\begin{abstract}

Despite their wide-spread success, Text-to-Image models (T2I) still struggle to produce images that are both aesthetically pleasing and faithful to the user's input text. 
We introduce \textbf{\ours}, a model-agnostic training algorithm by design that improves T2I models to be faithful to the text input.
\ours builds off a recent insight from TIFA's evaluation framework --- that large vision-language models (VLMs) can effectively identify the fine-grained discrepancies between generated images and the text inputs. 
\ours uses this insight to train T2I models without any labeled data; it improves T2I models using its own generations. 
First, it prompts the model to generate several candidate images for a given input text. Then, it uses two VLMs to select the best generation: a Visual Question Answering model that measures the alignment of generated images to the text, and another that measures the generation's aesthetic quality. After selection, we use LoRA to iteratively finetune the T2I model to guide its generation towards the selected best generations. \ours does not need any additional human annotation, model architecture changes, or reinforcement learning. Despite its simplicity, \ours improves both the semantic alignment and aesthetic appeal of two diffusion-based T2I models, evidenced by multiple benchmarks (+1.7\% on TIFA, +2.9\% on DSG1K, +3.4\% on VILA aesthetic) and human evaluation.
\end{abstract}
\vspace{-12pt} 

\begin{figure*}[thp]
    \centering
    \includegraphics[width=\linewidth]{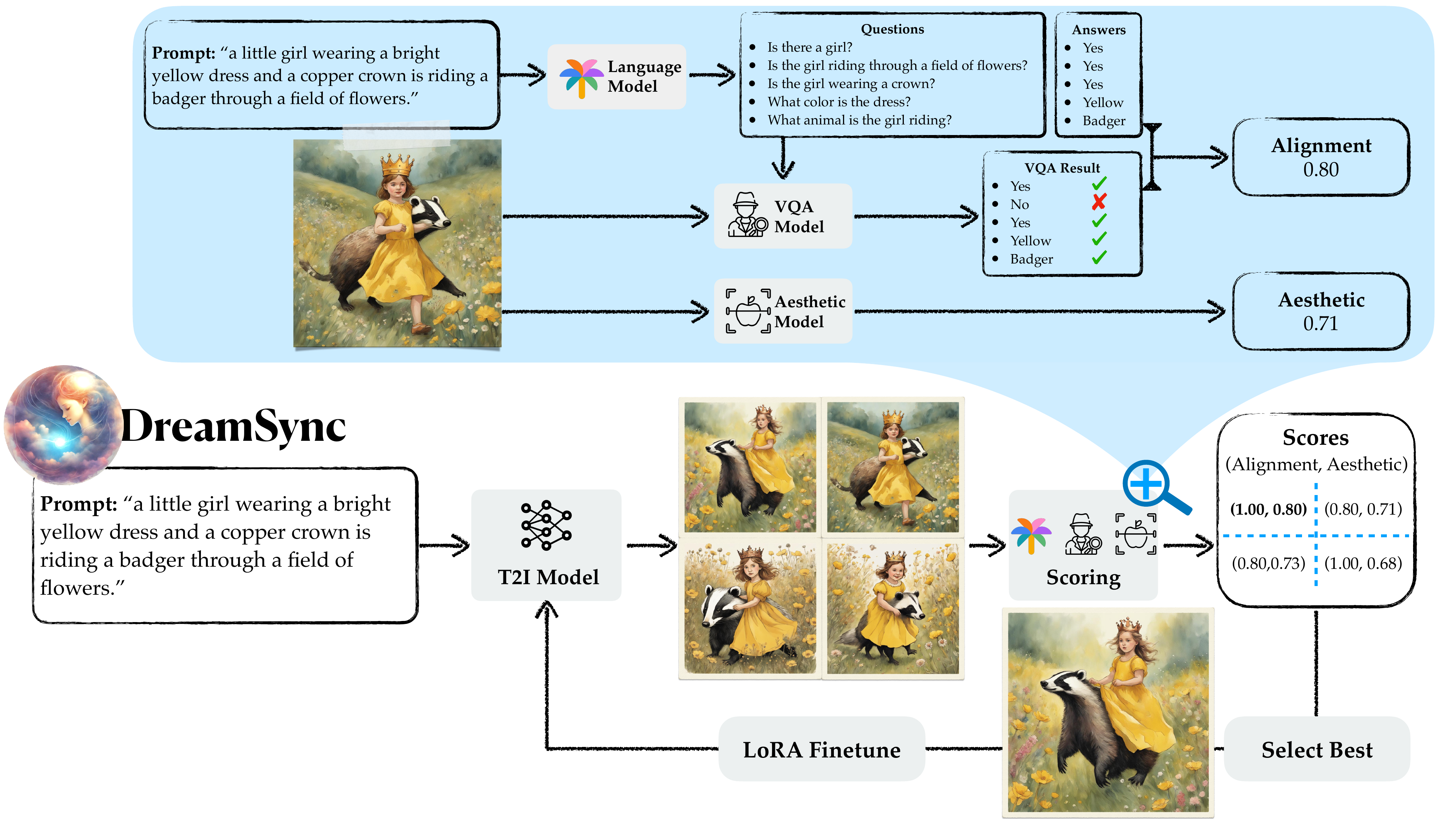}
    \caption{\textbf{\ours}. Given a prompt, a text-to-image generation model generates multiple candidate images, which are evaluated by two VLM models: one VQA model that provides feedback on text faithfulness and the other on image aesthetics. The best image chosen by the VLMs are collected to fine tune the T2I model. This process can repeat indefinitely until convergence on feedback is achieved. 
    }
    \label{fig:pipeline}
\end{figure*}

\section{Introduction}
\label{sec:intro}

Although we invite creative liberty when we commission art, we expect an artist to follow our instructions. 
Despite the advances in text-to-image (T2I) generation models~\cite{Ramesh2021ZeroShotTG, Rombach2021HighResolutionIS, Ramesh2022HierarchicalTI,Saharia2022PhotorealisticTD,Yu2022ScalingAM}, it remains challenging to obtain images that meticulously conform to users' intentions~\cite{petsiuk2022human, Feng2022TrainingFreeSD, lee2023aligning, Liu2022CompositionalVG, Liu2022CharacterAwareMI}. 
Current models often fail to compose multiple objects~\cite{petsiuk2022human, Feng2022TrainingFreeSD, Liu2022CompositionalVG}, bind attributes to the wrong objects~\cite{Feng2022TrainingFreeSD}, and struggle to generate visual text~\cite{Liu2022CharacterAwareMI}.
In fact, the difficulty of finding effective textual prompts has led to a myriad of  websites and forums dedicated to collecting and sharing useful prompts (e.g. PromptHero, Arthub.ai, Reddit/StableDiffusion). There are also online marketplaces for purchasing and selling useful such commands (e.g. PromptBase).
The onus to generate aesthetic images that are faithful to a user's desires should lie with the model and \textit{not} with the user.

Today, there are efforts to address these challenges.
For example, it is possible to manipulate attention maps based on linguistic structure to improve attribute-object binding~\cite{Feng2022TrainingFreeSD,royi}; or train reward models using human feedback to better align generations with user intent~\cite{lee2023aligning,fan2023dpok}.
Unfortunately, these methods either operate on a specific model architecture~\cite{Feng2022TrainingFreeSD,royi}
or require expensive labeled human data~\cite{lee2023aligning,fan2023dpok}. Worse, most of these methods sacrifice aesthetic appeal when optimizing for faithfulness, which we confirm in our experiments.

We introduce \textbf{\ours, a model-agnostic framework that improves T2I generation faithfulness while maintaining aesthetic appeal}.
Our approach extends work on fine-tuning T2I models for alignment, but does not require any human feedback.
The key insight behind \ours is in leveraging the advances in vision-language models (VLMs), which can identify fine-grained discrepencies between the generated image and the user's input text~\cite{hu2023tifa,dsg}.
Intuitively at a high level, our method can be thought of as a scalable version of reinforcement learning with human feedback (RLHF); just as LLaMA2~\cite{llama2} was iteratively refined using human feedback, \ours improves T2I models using feedback from VLMs, except without the need for reinforcement learning.

Given a set of textual prompts, T2I models first generates multiple candidate images per prompt.
\ours automatically evaluates these generated images using two VLMs.
The first one measures the generation's faithfulness to the text~\cite{hu2023tifa,dsg}, while the second one measures aesthetic quality~\cite{vila}.
The best generations are collected and used to finetune the T2I model using parameter-efficient LoRA finetuning~\cite{hu2022lora}.
With the new finetuned T2I model, we repeat the entire process for multiple iterations: generate images, curate a new finetuning set, and finetune again.

We conduct extensive experiments with latest benchmarks and human evaluation. 
We experiment \ours with two T2I models, SDXL~\cite{sdxl} and SD v1.4~\cite{sdv1.4}.
Results on both models show that \textbf{\ours enhance the alignment of images to user inputs and retains their aesthetic quality.} 
Specifically, quantitative results on TIFA~\cite{huang2023t2i} and DSG~\cite{dsg} demonstrate that \textbf{\ours is more effective than all baseline alignment methods} on SD v1.4, and can yield even bigger improvements on SDXL. Human evaluation on SDXL shows that \ours give consistent improvement on all categories of alignment in DSG.
While our study primarily focuses on boosting faithfulness and aesthetic quality, \ours has broader applications: it can be used to improve other characteristics of an image as long as there is an underlying model that can measure that characteristic.

\section{Related Work} \label{sec:related} 

\paragraph{T2I Evaluation with VLMs.} 
Several prior works have proposed to use VQA models to evaluate text-to-image generation. The TIFA benchmark, which pioneered this approach for evaluation, consists of 4K prompts and 25K questions across 12 categories (e.g., object, count, material), enabling T2I model evaluation by using VQA models to answer questions about the generated images~\citep{hu2023tifa}. TIFA prompts come from various resources, including DrawBench used in Imagen~\cite{Saharia2022PhotorealisticTD}, PartiPrompt used in Parti~\cite{Yu2022ScalingAM}, PaintSkill~\cite{Cho2022DALLEvalPT} used in Dall-Eval, etc. DSG~\cite{dsg} further improves TIFA's realiability by examining their evaluation questions carefully. Another related benchmark is SeeTrue, which also uses VQA models to measure alignment~\citep{yarom2023you}.
Before the VQA evaluation era, several other evaluation benchmarks were proposed focusing primarily on compositional text prompts for attribute binding (e.g., color, texture, shape) and object relationships (e.g., spatial).
Examples include T2I-CompBench~\citep{huang2023t2i}, C-Flowers~\cite{flower}, CC-500 and ABC-6K benchmarks~\cite{williamwang}. Aside from automated benchmarks, human evaluation for text-to-image generation is widely used in the community, although such annotations are notoriously costly to collect. In response,  \citet{xu2023imagereward} propose ImageReward, the first general purpose text-to-image human preference reward model to encode human preferences automatically. In our work, we use a collection of three evaluation methods to evaluate \ours: VQA evaluation for generated images on both TIFA and DSG benchmarks, human evaluation, and ImageReward for automatic human preference prediction. 

\paragraph{Improving General T2I Alignment.} We roughly categorize the alignment methods for improving T2I alignment into two classes depending on if they involve training. 
For training-involved methods, several works use Reinforcement Learning from Human Feedback (RLHF) based on human rankings to maximize a reward and improve faithful generation~\citep{fan2023dpok, karthik2023if, lee2023aligning}.   In a similar vein, Pick-a-Pic is a dataset of prompts and preferences that is used to train a CLIP-based scoring function~\citep{kirstain2023pick}. StyleDrop trains adapters to synthesize of images that follow a specific style~\citep{sohn2023styledrop}, and T2I-Adapter trains adapters to improve the control for the color and structure of the generation results~\citep{mou2023t2i}. DreamBooth and HyperDreamBooth improve personalized generation~\citep{ruiz2023dreambooth, ruiz2023hyperdreambooth}, and they have inspired more efficient methods such as SVDiff~\citep{han2023svdiff}. 
Being orthogonal to training-involved methods, there is a body of work on training-free methods that make inference time adjustments to the model to improve alignment, such such as SynGen and StructuralDiffusion.~\citep{hong2022improving, epstein2023diffusion, williamwang, royi}. \ours leverages training but does not involve reinforcement learning.  We compare \ours with two RL-based methods and two learning-free methods in our experiments. We find that \ours outperform all the baselines in terms of text-image alignment on both DSG and TIFA.

\paragraph{Iterative Bootstrapping.} Iterative Bootstrapping, also known as model self-training, is a semi-supervised learning approach that utilizes a teacher model to assign labels to unlabelled data, which is then used to train a student model \citep{yarowsky-1995-unsupervised,mcclosky-etal-2006-effective,pmlr-v119-kumar20c,fu-etal-2023-self-label}. In our work, we adopt a self-training scheme where the teacher model are the VLMs and the student model is the T2I model we aim to improve. During training, the VLMs (teacher) are used to annotate and select aligned examples for the next batch finetuning (student).

\section{\ours} \label{sec:method}
\begin{figure*}[tp]
    \centering
    \includegraphics[width=0.95\linewidth]{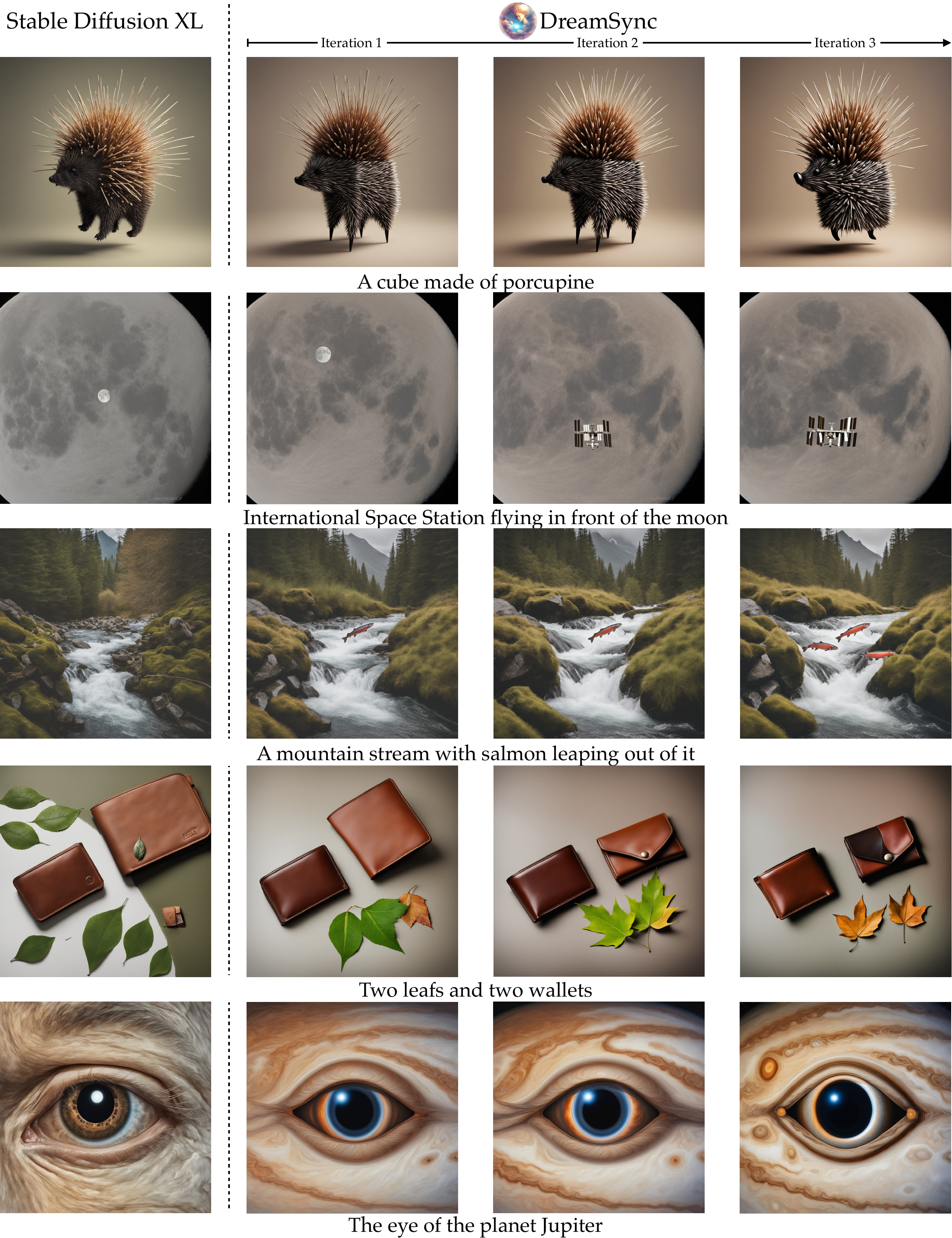}
    \caption{Qualitative examples of \ours improving image-text alignment after each iteration. LoRA fine-tuning on generated and filtered prompt-image pairs can steer the model to gradually capture more components of the text inputs. 
    }
    \label{fig:iterative_improvements}
\end{figure*}

Our method improves alignment and aesthetics in four steps (see Figure~\ref{fig:pipeline}): Sample, Evaluate, Filter, and Finetune. 
The high level idea is that T2I models are capable of generating interesting and varied samples. These examples are further judged by VLMs to pass qualification as faithful and aesthetic candidates for further finetuning T2I models. We next dive into each component more formally.

\paragraph{Sample.} Given a text prompt $T$, the text-to-image generation model $G$ generates an image $I = G(T)$. Generation models are randomized, and running $G$ multiple times on the same prompt $T$ can produce different images, which we index as $\{I^{(k)}\}_{k=1}^K$. To improve the model's faithfulness to text guidance, our method collects faithful examples generated by $G$. We use $G$ to generate $K$ samples of the same prompt $T$, so that with some probability $\delta > 0$, a generated image $I$ is faithful. Note that we need $K = \Omega(1/\delta)$ samples for each prompt $T$, and \ours is not expected to improve totally unaligned models (with $\delta \rightarrow 0$). Prior work~\citep{karthik2023if} estimates that 5--10 samples can yield a good image, and hence, $\delta$ can be thought of as roughly 0.1 to 0.2.

\paragraph{Evaluate.} For each text prompt $T$, we derive a set of $\mathcal N_T$ question-answer pairs $\{\mathcal Q(T), \mathcal A(T)\}$ that can be used to test whether a generated image $I$ is faithful to $T$. We use an LLM to generate these pairs, only using the  prompt $T$ as input (with no images). Typically $\mathcal N_T \approx 10$.
We use VQA models to evaluate the faithfulness of the generation model, 
$
    F_j(T, I) = \mathbbm{1} \{\textrm{VQA}(I, \mathcal Q_j(T)) = \mathcal A_j(T) \}, 
$
for $j \in \{1, \dots, \mathcal N_T\}$.
We measure the faithfulness of a caption-image pair $(T, I)$ given all questions and answers, using two metrics. Intuitively, we can average the number of correct answers, or we can be more strict, and only count an image as a success if all the answers are correct. Formally, the \textit{Mean} score is the expected success rate
$$
    \mathcal S_\mathrm{M}(T, I) = \frac{1}{\mathcal N_T} \sum_{j=1}^{\mathcal N_T} F_j(T, I),
    \label{eqn:mean_score}
$$
and the \textit{Absolute} score is the absolute success rate
$$
    \mathcal S_\mathrm{A}(T, I) = \prod_{j=1}^{\mathcal N_T} F_j(T, I).
    \label{eqn:abs_score}
$$

\paragraph{Filter.} We combine text faithfulness and visual appeal (given by $\mathcal V(\cdot)$) as rewards for filtering. For a text prompt $T$ and its corresponding synthetic image set $\{I_k\}_{k=1}^K$, we select samples that pass both VQA and aesthetic filters:
$$
    \begin{aligned}
        C(T) =  \{(T, I_k) : \mathcal S_\mathrm{M}(T, I_k) &\geq \theta_\mathrm{Faithful}, \\
    \mathcal V(I_k) &\geq \theta_\mathrm{Aesthetic}\}
    \end{aligned}.
$$
To avoid an imbalanced distribution where easy prompts have more samples, which could cause adversely affected image quality, we select one representative image (denoted as $\hat{I}_T$) having the highest visual appeal for each $T$:
$$
    (T, \hat{I}_T) = \mathop{\mathrm{argmax}}_{\mathcal V(I_k)} C(T).
$$
We apply this procedure to all text prompts in our finetuning prompt set $\{T_i\}_{i=1}^N$ with $T_i \sim \mathcal D$, where $\mathcal D$ is a prompt distribution. After filtering, we collect a subset of examples, $D(G) := \bigcup_{i \in \{j \mid C(T_j) \neq \varnothing \}} \{(T_i, \hat{I}_{T_i})\}$, that meet our aesthetic and faithfulness criteria.
Note that it is possible for $C(T_i)$ to be empty, and we empirically show what fraction of the training data is selected in \Cref{fig:chart_improvements}.
We ablate other aspects of the selection procedure in \S~\ref{ssec:ablation}.

\paragraph{Finetune.} After obtaining a new subset of faithful and aesthetic text-image pairs, we fine-tune our generative model $G$ on this set. 
We denote the generative model after $s$ iterations of \ours as $G_s$, such that $G_0$ denotes the baseline model.
To obtain $G_{s+1}$ we fine-tune on data generated by $G_s$ after applying our filtering procedure as outlined above.
We follow the same loss objective and fine-tuning dynamics as LoRA \cite{hu2022lora}. Let $\Theta(\cdot)$ denote all parameters of a model, then the hypothesis class at iteration $s$ is: 
$$
    \mathcal G_s = \left\{G \mid \mathrm{rank}\Big(\Theta(G) - \Theta(G_s)\Big) \leq R \right\}.
$$
where $R$ denotes the rank of weight updates and in practice we choose $R = 128$ to balance efficiency and image quality.
Overall, the iterative training procedure is as follows: 
\begin{equation}
    G_{s+1} = \mathop{\mathrm{argmin}}_{G \in \mathcal G_s} \frac{1}{|D(G_s)|} \sum_{(T_j, I_j) \in D(G_s)} \ell(G(T_j), I_j).
    \label{eqn:iterative}
\end{equation}
The self-training process \cref{eqn:iterative} can in principle be executed indefinitely. In practice, it repeats for three iterations at which point we observe diminishing returns. 
 
\begin{figure}[t]
    \centering
    \includegraphics[width=0.9\linewidth]{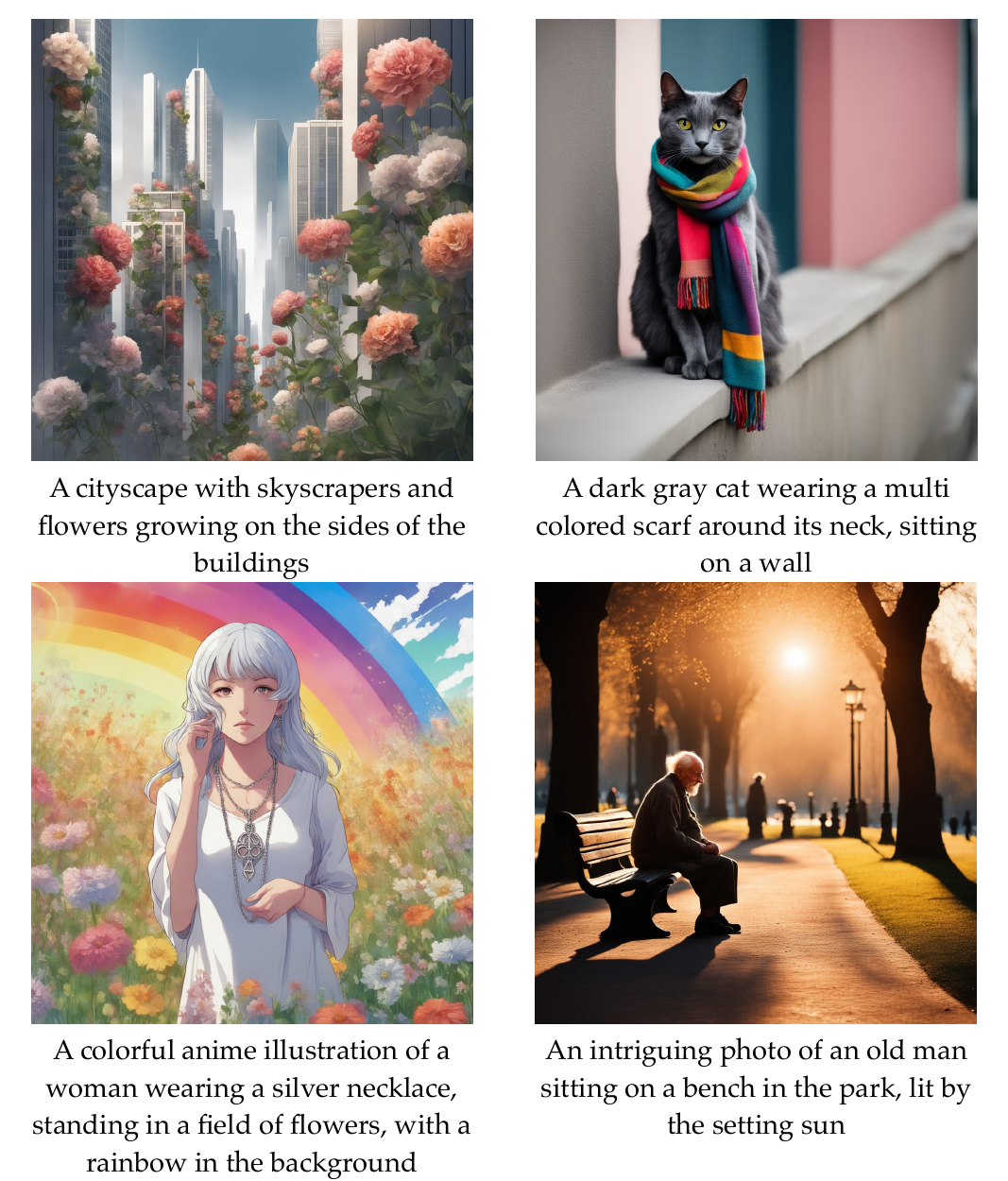}
    \caption{PaLM-2 generated training prompts and their corresponding images generated via \ours. Prompt acquisition requires no human effort. It enables us to train on more complex and diversified prompt-image pairs than found in typical datasets.
    }
    \label{fig:train_set}
\end{figure}

\begin{table*}[t]
    \centering
    \begin{tabular}{ll l cc   c  c}
    \toprule
       \multirow{3}*{\makecell{\textbf{Model}}} & & \multirow{3}*{\makecell{\textbf{Alignment}}} & \multicolumn{3}{c}{\textbf{Text Faithfulness}} & \multirow{3}*{\makecell{\textbf{Visual Appeal}}}  \\
        \cmidrule{4-6}
         & &  &  \multicolumn{2}{c}{\texttt{TIFA}}  & \multirow{2}*{\texttt{DSG1K}} & \\
         \cmidrule{4-5}
         & &  & \emph{Mean} & \emph{Absolute}  & \\
    \midrule
    \multirow{6}*{SD v1.4~\cite{sdv1.4}} &  & No alignment & \score{76.6}{} & \score{33.6}{} & \score{72.0}{} & \score{44.6}{}\\
    \cmidrule{2-3}
    & \multicolumn{1}{c}{\multirow{2}{*}{Training-Free}} & SynGen \citep{royi} & \scoreg{76.8}{+0.2} & \scoreg{34.1}{+0.5} & \scorel{71.2}{--0.8} & \scorel{42.4}{--2.2}\\
    & \multicolumn{1}{l}{} & StructureDiffusion \cite{williamwang} &  \scorel{76.5}{--0.1} & \scoreg{33.6}{+0.0} & \scorel{71.9}{--0.1} & \scorel{41.5}{--3.1} \\
    \cmidrule{2-3}
    & \multicolumn{1}{c}{\multirow{2}{*}{RL}} & DPOK \citep{fan2023dpok} &  \scorel{76.4}{--0.2} & \scoreg{33.8}{+0.2} & \scorel{70.3}{--1.7}  & \scoreg{\best{46.5}}{+1.9} \\ 
    & \multicolumn{1}{l}{} & DDPO \cite{black2023training} &  \scoreg{76.7}{+0.1} & \scoreg{34.4}{+0.8} & \scorel{70.0}{--2.0} & \scorel{43.5}{--1.1} \\
    \cmidrule{2-3}
    & & \dreamicon\xspace\ours\ (ours) & \scoreg{\best{77.6}}{+1.0} & \scoreg{\best{35.3}}{+1.7} & \scoreg{\best{73.2}}{+1.2} & \scoreg{44.9}{+0.3}\\
        \midrule\midrule
\multirow{2}*{SDXL~\cite{sdxl}} & & No alignment   &  \score{83.5}{} & \score{45.5}{} & \score{83.4}{} & \score{60.9}{}\\ 
   & & \dreamicon\xspace\ours\ (ours) & \scoreg{\best{85.2}}{+1.7} & \scoreg{\best{49.2}}{+3.7} & \scoreg{\best{86.3}}{+2.9} & \scoreg{\best{64.3}}{+3.4}\\
    \bottomrule
    \end{tabular}
    \caption{\textbf{Benchmark on Text Faithfulness and Visual Appeal.} All models are sampled with the same set of four seeds, i.e. $K=4$. Best scores under each backbone T2I model are highlighted in \best{bold}; {\color{ggreen}gain} and {\color{gred}loss} compared to base models are highlighted accordingly. \ours significantly improve SD-XL and SD v1.4 in alignment and visual appeal across all benchmark. Additionally, \ours does not sacrifice image quality when improving faithfulness.
    \vspace{-.1in}
    }
    \label{tab:main_result}
\end{table*}

\section{Datasets and Evaluation} 
\label{sec:data}
In this section, we will introduce our training data in \S~\ref{ssec:train_data} and evaluation benchmark in \S~\ref{ssec:eval_data}.

\subsection{Training Data Acquisition} 
\label{ssec:train_data}
To obtain prompts, and corresponding question-answer pairs without human-in-the-loop, we utilize the in-context learning capability of Large Language Models (LLM).
We choose PaLM 2\footnote{\url{https://ai.google/discover/palm2/}} \cite{anil2023palm} as our LLM and proceed as follows:

\begin{enumerate}
\item \emph{Prompt Generation.} We provide five hand-crafted seed prompts as examples and then ask PaLM 2 to generate similar textual prompts. We include additional instructions that specify the prompt length, a category (randomly drawn from twelve desired categories as in \cite{hu2023tifa}, e.g., spatial, counting, food, animal/human, activity), no repetition, etc.\footnote{In \Cref{appendix:instructions}, we show the complete instruction used to probe LLM for the first two steps: prompt generation and QA generation.} We change the seed prompts and repeat the prompt generation three times.

\item \emph{QA Generation.} Given prompts, we then use PaLM 2 again to generate question and answer pairs that we will use as input for VQA models as in TIFA~\cite{hu2023tifa}.

\item \emph{Filtering.} We finally use PaLM 2 once more to filter out unanswerable QA pairs. Here our instruction aims to identify three scenarios: the question has multiple answers (e.g., ``\texttt{black and white panda}'' where the object has multiple colors, each color could be the answer), the answer is ambiguous (e.g., ``\texttt{a lot of people}'') or the answer is not valid to the question. 
\end{enumerate}

We showcase the diversity of PaLM 2 generated prompts in \Cref{fig:train_set} using qualitative examples and  quantitive statistics of our generated prompts in \Cref{appendix:palm_statistics}.

\subsection{Evaluation Benchmarks}
\label{ssec:eval_data}
Using the previously generated prompts, we evaluate whether \ours can improve the T2I model performance on benchmarks that include general prompts. We consider the follow benchmarks.

\paragraph{TIFA.} To evaluate the faithfulness of the generated images to the textual input, TIFA~\cite{hu2023tifa} uses VQA models to check whether, given a generated image, questions about its content are answered correctly. There are 4k diverse prompts and 25k questions spread across 12 categories in the TIFA benchmark. Although there is no overlap between our training data and TIFA, we use the TIFA attributes to constrain our LLM-based prompt generation. Therefore, we use TIFA to test \ours on in-distribution prompts. We follow TIFA and use BLIP-2 as the VQA model for evaluation.

\paragraph{Davidsonian Scene Graph (DSG).} 
DSG~\cite{dsg} exhibits the same VQA-as-evaluator insight as TIFA's and further improves its reliability. Specifically, DSG ensures that all questions are atomic, distinct, unambiguous, and valid. To comprehensively evaluate T2I images, DSG provides 1,060 prompts covering many concepts and writing styles from different datasets that are completely independent from \ours's training data acquisition stage. Not only is DSG a strong T2I benchmark, it also enables further analysis of \ours with out-of-distribution prompts. Furthermore, DSG uses PaLI as the VQA model for evaluation, which is different from the VQA model that we use in training (\emph{i.e.}, BLIP-2) and lifts the concern of VQA model bias in evaluation. We use DSG QA both automatically (with PaLI) and with human raters (details in \Cref{dsg-appendix}).

\section{Experiments} 
\label{sec:experiments}
We explain our experimental setup in \S~\ref{ssec:model}, and showcase the efficacy of training with \ours and compare against other methods in \S~\ref{ssec:main_result}.
\S~\ref{ssec:ablation} analyzes our choice of rewards; \S~\ref{ssec:human_eval} reports results for a human study.

\subsection{Experimental set-up}
\label{ssec:model}
\paragraph{Base Model.} We evaluate \ours on Stable Diffusion v1.4~\citep{sdv1.4}, which is also used in related work.
Additionally, we consider SDXL~\citep{sdxl}, which is the current state-of-the-art open-sourced T2I model. For each prompt, we generate eight images per prompt, i.e., $K=8$.

\paragraph{Fine-grained VLM Feedback.} We use feedback from two VLM models to decide what text-image pairs to keep for finetuning. We use BLIP-2~\cite{Li2023BLIP2BL} as the VQA model to measure the faithfulness of generated images to textual input and and VILA~\cite{vila} to measure the aesthetics measurement score. Empirically, we keep the text-image pairs whose VQA scores are greater than $\theta_\mathrm{Faithful}=0.9$ and aesthetics score greater than $\theta_\mathrm{Aesthatics}=0.6$.
If there are multiple generated images passing the threshold, we keep the one with the highest VILA score.  
Starting from 28,250 prompts, we find that more than 25\% prompts are kept for $D(G_0)$ (for both T2I models), which we will use for finetuning. We later show that this percentage increases further as we perform additional \ours iterations.

\paragraph{Baselines.} 
We compare \ours with two types of methods that improve the faithfulness of T2I models: two training-free methods (StructureDiffusion~\cite{williamwang} and SynGen~\cite{royi}) and two RL-based methods (DPOK~\cite{fan2023dpok} and DDPO~\cite{black2023training}).  
As the baselines use SD v1.4 as their backbone, we also use it with \ours for a fair comparison. 

\begin{figure}[t]
    \centering
    \includegraphics[width=\linewidth]{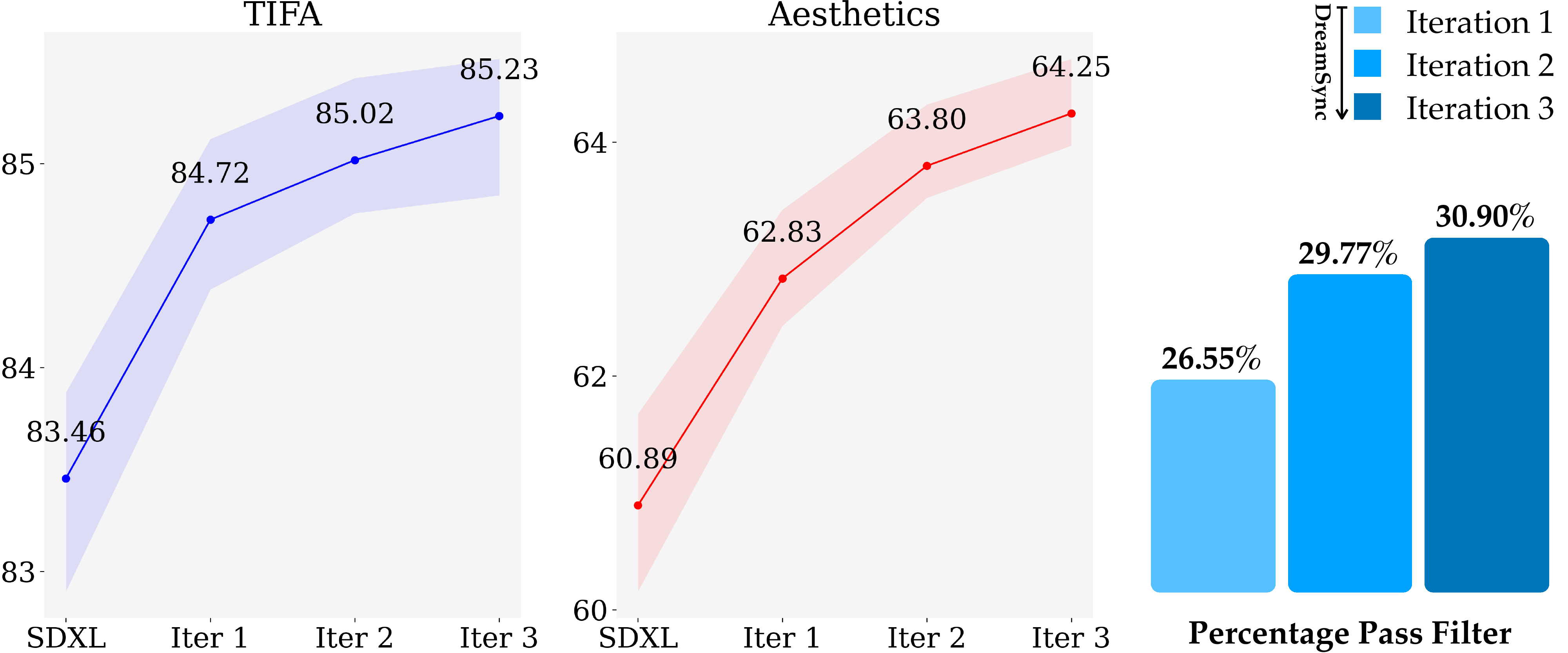}
    \caption{\ours improves faithfulness and aesthetics iteratively. More examples pass the filters with additional iterations.\vspace{-.1in}}
    \label{fig:chart_improvements}
\end{figure}

\subsection{Benchmark Results}
\label{ssec:main_result}
In \Cref{tab:main_result} we compare \ours to various state-of-the-art approaches with four random seeds. In \Cref{appendix:qualitative_sdv1_4,appendix:qualitative_sdxl} we show more qualitative comparisons.

\paragraph{\ours Improves the Alignment and Aesthetics of both SDXL and SD v1.4.} For SDXL~\cite{sdxl}, we show how three iterations of \ours improves the generation faithfulness by 1.7 point of mean score and 3.7 point of absolute score on TIFA.
The visual aesthetic scores after performing \ours improved by 3.4 points.
Due to the model-agnostic nature, it is straightforward to apply \ours to different T2I models.
We also apply \ours to SD V1.4~\cite{sdv1.4}.
\ours improves faithfulness by 1.0 points of mean score and 1.7 points of absolute score on TIFA, together with a 0.3 points of VILA score improvement for aesthetics.
Most prominently on DSG1K, \ours improve text faithfulness of SDXL by 2.9 points. We report fine-grained results for DSG in Appendix \ref{dsg-appendix}. 

\paragraph{\ours yields the best performance in terms of textual faithfulness on TIFA and DSG.} This is true without sacrificing the visual appearance as shown in \Cref{tab:main_result}.
In \Cref{fig:chart_improvements} we report TIFA and aesthetics scores for each iteration, where we observe how \ours gradually improves the alignment and aesthetics of the generated images. 
We highlight several qualitative examples in \Cref{fig:iterative_improvements}.

\subsection{Analysis \& Ablations}
\begin{table}[t]
\footnotesize
    \centering
    \begin{tabular}{cc p{0.01cm} c p{0.01cm} c}
    \toprule
  \multicolumn{2}{c}{Rewards} && \multirow{2}*{\makecell{Text\\Faithfulness}} && \multirow{2}*{\makecell{Visual\\Appeal}} \\
        \cmidrule{1-2}
     VQA     & VILA   &&    &&  \\
    \midrule
    -  &  - &&  \score{83.5}{0.39}   &&  \score{60.9}{0.56}\\
    \midrule
    \checkmark  &   &&  \score{\textbf{84.8}}{0.48}   &&  \score{61.9}{0.44}\\   
    &   \checkmark  && \score{83.8}{0.30}   &&  \score{61.7}{0.24}\\ 
     \checkmark &  \checkmark && \score{84.7}{0.29}  &&  \score{\textbf{62.8}}{0.37}\\ 
    \bottomrule
    \end{tabular}
    \caption{Ablation of different VLM rewards. Models are evaluated after \textit{one \ours iteration}.\vspace{-.1in}}
    \label{tab:ablation}
\end{table}
\label{ssec:ablation}

\paragraph{Impact of VQA model on evaluation.}
We analyze whether using BLIP-2 as a VQA model for finetuning and for evaluation in TIFA might be the reason for the improvement by \ours that we have observed.
To test this we use PaLI~\cite{PaLI} to replace the BLIP-2 as the VQA in TIFA.
Using SDXL as the backbone, \ours improves the mean score from 90.09 to 92.02 on TIFA compared to the vanilla SDXL model.
This results confirms that \ours is in fact able to improve the textual faithfulness of T2I models.

\begin{table}[t]
\footnotesize
    \centering
    \begin{tabular}{l  l  c  c }
\toprule
  \multirow{2}*{\makecell{T2I\\Model}} & \multirow{2}*{\makecell{Alignment\\Method}} & \multicolumn{2}{c}{Evaluation Dataset} \\
  \cmidrule{3-4}
  & &  TIFA & DSG1K \\
  \midrule
  \multirow{4}*{SD v1.4} & No alignment & 0.056 &  -0.220 \\
             & SynGen & 0.149 & -0.237 \\
              & StructureDiffusion & 0.075 &  -0.135\\
              & DPOK & 0.067  & -0.258\\
              & DDPO & 0.152 & -0.076\\
              & \ours (ours) & \best{0.168}  & \best{-0.054} \\ 
  \midrule
  \multirow{2}*{SD XL} & No alignment & 0.878  & 0.702\\
            & \ours (ours) & \best{1.020} & \best{0.837}\\
\bottomrule
    \end{tabular}
    \caption{Scores given by the human preference model ImageReward~\cite{xu2023imagereward}; model scores are logits and can be negative. Models trained with \ours outperform other baselines (higher is better), without using any human annotation.\vspace{-.1in}}
    \label{tab:image_rewards}
\end{table}

\paragraph{Ablating the Reward Models} 
In \Cref{tab:ablation}, we present the results for an ablation study where we remove one of the VLMs during filtering and evaluate SDXL after applying one iteration of \ours. 
It can be seen how training with a single pillar mainly leads to an improvement in the corresponding metric, while the combination of the two VLM models leads to strong performance for both text faithfulness and visual easthatics, justifying our approach. 
One interesting finding is that training with both rewards, rather than VILA only, gives the highest visual appeal score. Our possible explanation is that images that align with user inputs may have higher visual appeal.

\paragraph{ImageReward.} We next test whether \ours yields an improvement on human preference reward models, even though \ours is not trained to optimize them. We use ImageReward~\citep{xu2023imagereward} as an off-the-shelf human preference model for generated images.
\Cref{tab:image_rewards} shows that \ours plus either SD v1.4 or SDXL increases ImageReward scores on images based on both TIFA and DSG1K.
Tuning with VLM-based feedback helps align the generated images with human preferences, at least according to ImageReward.

\begin{figure}[t]
    \centering
    \includegraphics[width=\linewidth]{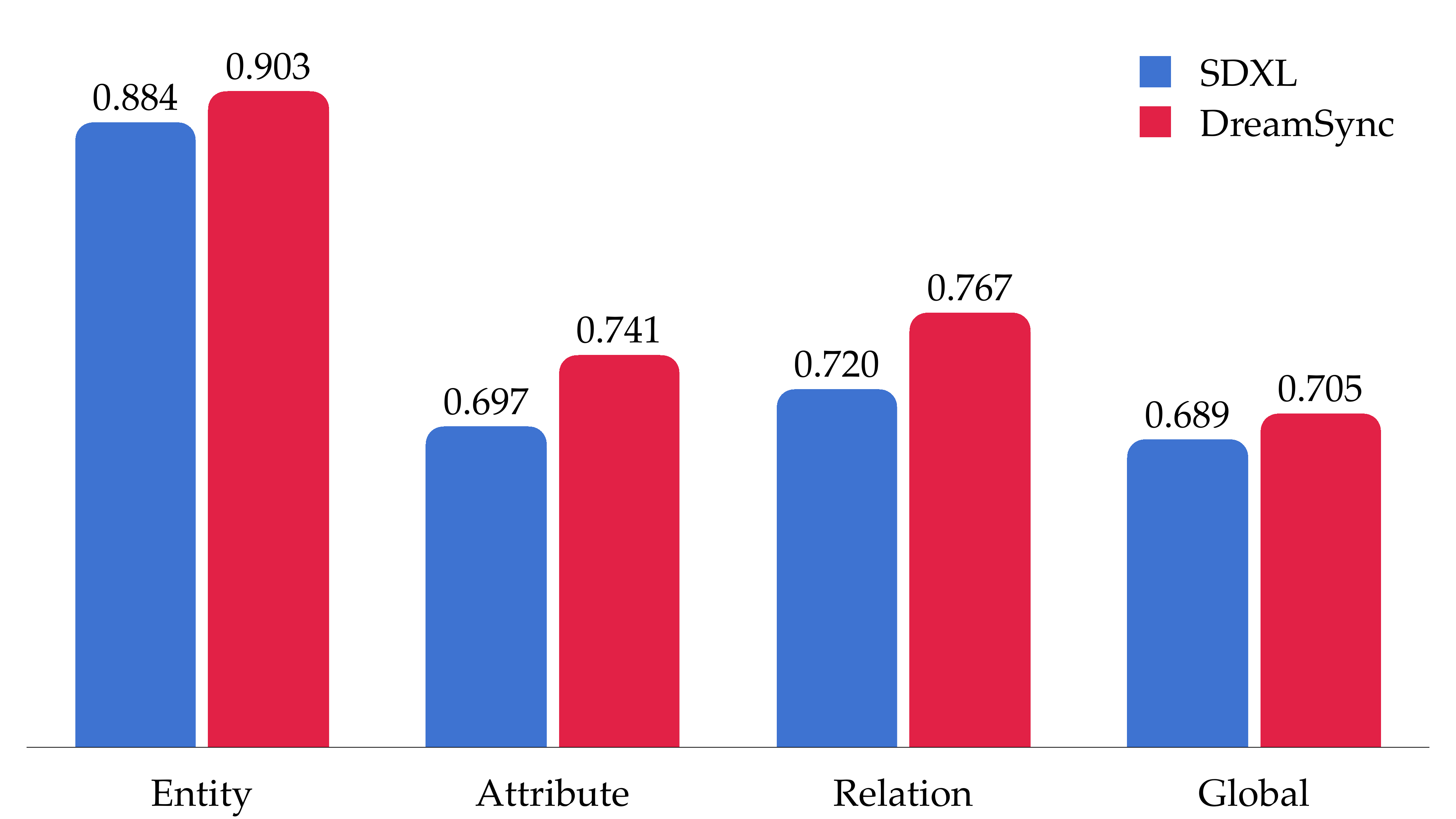}
    \caption{Human study with three raters on 1060 DSG prompts.}
    \label{fig:dsg-human-broad}
\end{figure}

\subsection{Human Evaluation} \label{ssec:human_eval}
To corroborate the VQA-based results, we first conduct a preliminary human study to evaluate the faithfulness of generated images. It shows simply asking one question \texttt{``Which image better aligns with the prompt?''} yields poor inter-annotator agreement. We speculate that asking a single question encompassing the whole prompt makes the alignment difficult to evaluate.

To address this issue, we conduct a larger follow-up study based on DSG~\cite{dsg}, where we ask approximately 8 fine-grained questions for each of 1060 images to external raters. These questions are divided into categories (entity, attribute, relation, global). Here in \Cref{fig:dsg-human-broad}, we observe consistent and statistically significant  improvements comparing \ours to SDXL. In each category, images from \ours contain more components of the prompts, while excluding extraneous features. Overall, \ours's images led to 3.4\% more correct answers than SDXL images, from 70.9\% to 74.3\%. Full details and findings for both studies are in Appendix \ref{dsg-appendix}.

\section{Discussion}

A key design choice behind \ours is to maintain simplicity and automation throughout each step of the pipeline. Despite this feature, our experimental results show that \ours can improve both SD v1.4 and SDXL on TIFA, DSG, and visual appeal. In the case of SD v1.4, this improvement holds true compared four different baseline models (two training-free and two RL-based). For SDXL, even though the base model achieves SoTA results among open-source models, \ours can still substantially improve both alignment and aesthetics.

The effectiveness of \ours's self-training methodology opens the door for a new paradigm of parameter-efficient finetuning. Indeed, the \ours pipeline is easily generalizable. For the training prompts, we can construct a set with complex and non-conventional examples compared to standard web-scraped data. On the filtering and fine-tuning side, our framework shows that VLMs can provide effective feedback for T2I models. Together, these steps do not require human annotations, yet they can tailor a generative model toward desirable criteria.

\subsection{Limitations}
\label{ssec:limitations}
Like prior methods, the performance of \ours is limited by the pre-trained model it starts with. As exemplified in ``the eye of the planet Jupiter'' in Figure~\ref{fig:iterative_improvements}, SDXL generates a human's eye rather than Jupiter's. \ours adds more features of the Jupiter in each iteration. Nevertheless, it did not manage to produce an image that is perfectly faithful to the prompt. This is also exemplified by the quantitative results in \S\ref{ssec:main_result}. Despite outperforming the baselines using SD v1.4 on TIFA and DSG, SD v1.4 + \ours still falls behind SDXL. Similarly,  our human studies on DSG in \S\ref{ssec:human_eval} indicate that \ours improves SDXL from 70.9\% accuracy to 74.3\%. Nonetheless, there is still a 25.7\% headroom to improve.  We identify several common failure modes (e.g., attribute-binding) and conduct a detailed analysis in Appendix~\ref{appendix:more-examples}. Future works may investigate if these challenges can be addressed by further scaling up \ours, or mixing it with large-scale pre-training.

\section{Conclusion}\label{sec:conclusion}
We introduce \ours, a versatile framework to improve text-to-image (T2I) synthesis with feedback from image understanding models. Our dual VLM feedback mechanism helps in both the alignment of images with textual input and the aesthetic quality of the generated images. Through evaluations on two challenging T2I benchmarks (with over five thousand prompts),  we demonstrate that \ours can improve both SD v1.4 and SDXL for both alignment and visual appeal. The benchmarks also show that \ours performs well in both in-distribution and out-of-distributions settings. Furthermore, human ratings and a human preference prediction model largely agree with \ours's improvement on benchmark datasets.

For future work, one direction is to ground the feedback mechanism to give fine-grained annotations (e.g., bounding boxes to point out where in the image the misalignment lies).
Another direction is to tailor the prompts used at each iteration of \ours to target different improvements: backpropagating VLM feedbacks to the prompt acquisition pipelines for continual learning. 

\clearpage
\iftrue

\section*{List of Contributions}
{
\setlength\parindent{0pt}
\textbf{Jiao Sun}: Jiao leads the project. She implemented \ours internally at Google, showcasing its success on various reward models. She also wrote the first paper draft.

\vspace{0.5ex}
\textbf{Deqing Fu}: Deqing initiated the idea of LLM-based prompt generation. He implemented \ours with open-source text-to-image models, conducting all the experiments. He also contributed to paper writing and drew all the figures.

\vspace{0.5ex}
\textbf{Yushi Hu}: Yushi conceived the idea of \ours and designed the experiments. He also implemented the VQA feedback, the baselines, and contributed to paper writing.

\vspace{0.5ex}
\textbf{Jiao, Deqing, and Yushi} completed the full development cycle of \ours. They contributed equally on designing technical directions, implementing, polishing and improving \ours from scratch. 

\vspace{0.5ex}
\textbf{S.~Wang} ran both automatic and human evaluations on \texttt{DSG1K} and contributed to corresponding paper sections. 

\vspace{0.5ex}
\textbf{R.~Rassin} contributed to the implementation of the baseline methods SynGen and DPOK. 

\vspace{0.5ex}
\textbf{J.~Sun},
\textbf{C.~Rashtchian}, \textbf{S.~van Steenkiste}, \textbf{C.~Herrmann}, \textbf{D-C.~Juan}, and \textbf{D.~Alon} conceived of the initial project directions, e.g., T2I models struggle with compositionality and image quality. 

\vspace{0.5ex}
\textbf{R. Krishna}, \textbf{S.~van Steenkiste}, \textbf{C.~Herrmann}, and \textbf{D.~Alon} provided constructive feedback and suggested experiments.

\vspace{0.5ex}
\textbf{R.~Krishna} and \textbf{S.~van Steenkiste} helped frame the story via writing and polishing several sections of the paper.

\vspace{0.5ex}
\textbf{C.~Rashtchian} served as senior project lead and manager, scoping technical directions and facilitating collaborations.

\subsection*{Acknowledgments} We thank Yi-Ting Chen, Otilia Stretcu, Yonatan Bitton, Fei Sha, Kihyuk Sohn, Chun-Sung Ferng, and Jason Baldridge for helpful project discussions and technical support. JS and DF would like to thank USC NLP group and YH would like to thank UW NLP group, for providing both additional GPU computational resources and fruitful discussions.

}

\fi

{
\newpage
\small
\bibliographystyle{ieeenat_fullname}
\bibliography{main}
}

\clearpage
\onecolumn
\appendix
\section{Training Data Acquisition}
\label{appendix:train_data}

\subsection{LLM Instructions}
\label{appendix:instructions}
Training Data Acquisition is the first step and the foundation of \ours as discussed in \Cref{ssec:train_data}. We use PaLM 2 for each step of the training data acquisition, including prompt generation, QA generation and filtering. Here are the complete instructions that we use.

\paragraph{Instruction for Prompt Generation.}  \texttt{You are a large language model, trained on a massive dataset of text. You can generate texts from given examples.
You are asked to generate similar examples to the provided ones and follow these rules:
\begin{enumerate}
    \item Your generation will be served as prompts for Text-to-Image models. So your prompt should be as visual as possible.
    \item Do NOT generate scary prompts.
    \item Do NOT repeat any existing examples.
    \item Your generated examples should be as creative as possible.
    \item Your generated examples should not have repetition.
    \item Your generated examples should be as diverse as possible.
    \item Do NOT include extra texts such as greetings.
\end{enumerate}}

\paragraph{Instruction for  QA Generation.} \texttt{Given a image descriptions, generate one or two multiple-choice questions that verifies if the image description is correct.
Classify each concept into a type (object, human, animal, food, activity, attribute, counting, color, material, spatial, location, shape, other), and then generate a question for each type.} We then provide fifteen prompts together with about ten question answer pairs as demonstration for PaLM 2. \Cref{tab:qa_example} shows an example of PaLM2-generated \emph{prompt} and \emph{QA}. \emph{Answer source} and \emph{Answer Type} are also automatically generated altogether, making it possible for us to get statistics of our training set below.

\begin{table}[b]
\centering
\ttfamily
\footnotesize
\begin{tabular}{@{}l|l|l|l@{}}
\toprule
\textbf{Prompt} & \textbf{Question and Choices} & \textbf{Answer Source} & \textbf{Answer Type} \\ \midrule
\multirow{9}{*}{\begin{tabular}[c]{@{}l@{}}6 baseball players, \\ each holding a sheep, \\ and they are all standing\\  in a field of flowers\end{tabular}} & \begin{tabular}[c]{@{}l@{}}question : what is in the field?  \\ choices: {[}"\textbf{flowers}", "grass", "trees", "rocks"{]}\end{tabular} & flowers & object \\ \cmidrule(l){2-4} 
 & \begin{tabular}[c]{@{}l@{}}is there a field?\\ choices: {[}"\textbf{yes}", "no"{]}\end{tabular} & field & location \\ \cmidrule(l){2-4} 
 & \begin{tabular}[c]{@{}l@{}}are there flowers?\\ choices: {[}"\textbf{yes}", "no"{]}\end{tabular} & flowers & object \\ \cmidrule(l){2-4} 
 & \begin{tabular}[c]{@{}l@{}}what type of place is this?\\ choices: {[}"\textbf{field}", "park", "forest", "mountain"{]}\end{tabular} & field & location \\ \cmidrule(l){2-4} 
 & \begin{tabular}[c]{@{}l@{}}are the baseball players holding sheep?\\ choices: {[}"\textbf{yes}", "no"{]}\end{tabular} & holding & activity \\ \cmidrule(l){2-4} 
 & \begin{tabular}[c]{@{}l@{}}are there sheep?\\ choices: {[}"\textbf{yes}", "no"{]}\end{tabular} & sheep & animal \\ \cmidrule(l){2-4} 
 & \begin{tabular}[c]{@{}l@{}}are there baseball players?\\ choices: {[}"\textbf{yes}", "no"{]}\end{tabular} & baseball players & human \\ \cmidrule(l){2-4} 
 & \begin{tabular}[c]{@{}l@{}}how many baseball players are there?\\ choices: {[}"1", "2", "3", "4", "5", "\textbf{6}"{]}\end{tabular} & 6 & human \\ \cmidrule(l){2-4} 
 & \begin{tabular}[c]{@{}l@{}}how many sheep are there?\\ choices: {[}"1", "2", "3", "4", "5", "\textbf{6}"{]}\end{tabular} & 6 & animal \\ \bottomrule
\end{tabular}
\caption{One example of PaLM2-generated \emph{prompt} and \emph{QA}. \emph{Answer source} and \emph{Answer Type} are also generated by PaLM 2, making it possible for us to get statistics of our training set. We highlight correct answers in \textbf{bold} here.}
\label{tab:qa_example}
\end{table}

\subsection{Statistics}
\label{appendix:palm_statistics}
\begin{table}[ht]
\centering
\begin{tabular}{lr}
\toprule[1.2pt]
\# of prompts & 28,250 \\
\midrule
\# of questions & 239,310 \\
$\ $ - \# of binary questions & 125,094 \\
$\ $ - \# of multiple-choice questions & 114,214 \\
\midrule
avg. \# of questions per prompt & 8.5 \\
avg. \# of words per prompt & 16.7 \\
avg. \# of elements per prompt & 1.9 \\
\bottomrule[1.2pt]
\end{tabular}
\caption{
Statistics of Training Set for \ours.
}
\label{tab:statistics}
\end{table}
Table~\ref{tab:statistics} shows the statistics of the prompts and questions we obtained, and we list a few prompts from our training set and \ours's generation in \Cref{fig:train_set}. 
Prior work (e.g., TIFA, DSG) identifies that T2I models do not perform equally well for depicting different attribute categories; we verify the variety of attributes in our prompts by counting unique words (i.e., \emph{Answer Source} in \Cref{tab:qa_example}) in these categories (i.e., \emph{Answer Type} in \Cref{tab:qa_example}):
counting (4179), object (3638), shape (973), human (945), location (1047), activity (2984), attribute (2925), color (3259), food (1086), spatial (1009), animal (645), material (1610), existence (3072), and other (878).

\clearpage
\subsection{Images Generated by \ours for Finetuning Exhibit High Quality} \label{app:more_images}

\begin{figure}[b]
    \centering
    \includegraphics[width=0.96\linewidth]{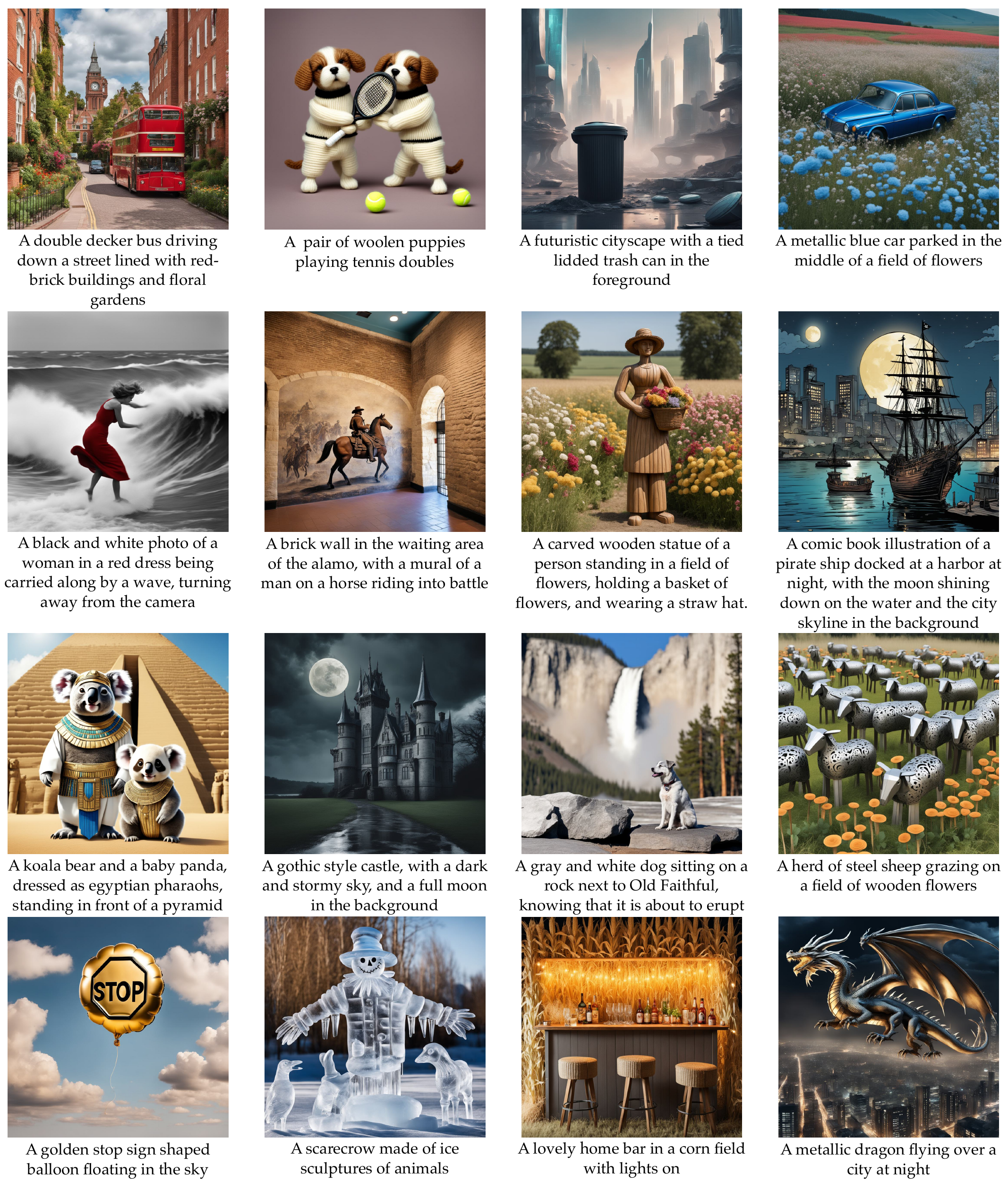}
    \caption{Prompts and Images Generated via \ours for Finetuning. Prompts generated by PaLM-2 exhibit high diversity and corresponding synthetic images exhibit high quality and alignment.}
    \label{fig:more_imgaes}
\end{figure}

\clearpage
\section{Failure Modes Analysis}
\label{appendix:more-examples}
\begin{figure}[!ht]
    \centering
    \includegraphics[width=0.99\linewidth]{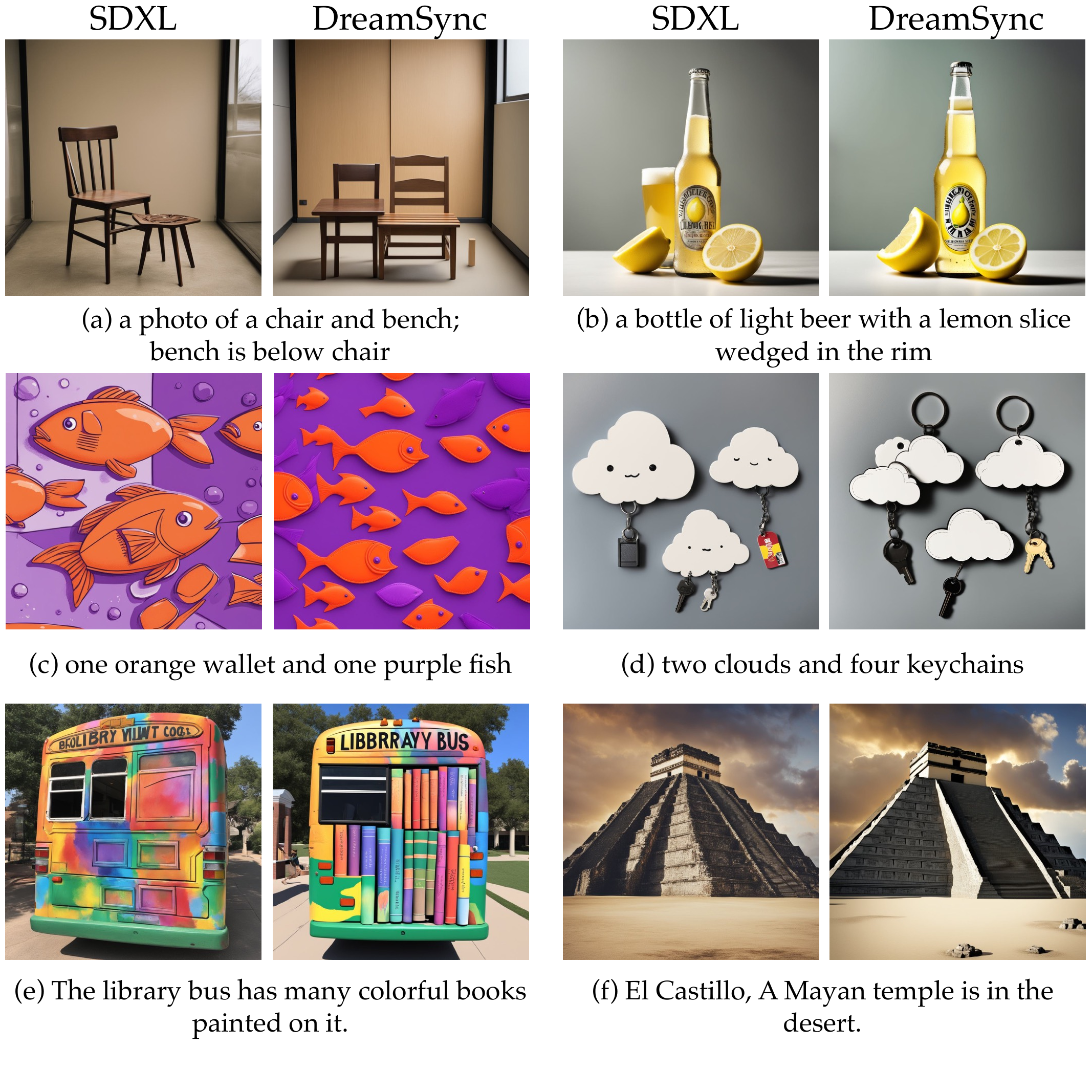}
    \caption{\textbf{Failure modes.} 
    We present qualitative examples of \ours failures. First, \textbf{it remains challenging to compose multiple objects and bind the attributes correctly}, as shown in (a), (b), (c), and (d). 
    Second, we observe that \textbf{the quality of details and shadows decline on some images}, as illustrated in (e) and (f). Overall, SDXL + \ours still has room for improvement in terms of text-to-image faithfulness and quality.
    }
    \label{fig:failure_modes_analysis}
\end{figure}

Figure~\ref{fig:failure_modes_analysis} presents several side-by-side examples showcasing common failure modes of \ours. For each example, we show the image generated by SDXL on the left, and the image of SDXL + \ours on the right. We also indicate some key directions for improvements.

\begin{itemize}
    \item \textbf{Composing multiple objects and attributes is still challenging.} As shown in (a), (b), (c), and (d), SDXL + \ours struggles to produce an image that is faithful to the prompt. In (a), \ours adds a bench in the image. However, the attributes of chairs and benches are mixed. In (b), \ours removes the extra glass in the background, but neither model is able to place the lemon wedge in the rim of the bottom.  In (c), \ours adds purple fish in the image, but the counting is not correct. In (d), \ours produces four objects but they are cloud-keychain combinations.
    \item \textbf{We observe decline of texture details and shadows on some images.} In (e), the alignment between the text and the bus significantly improves. 
    However, the quality of the bus shadow declines. 
    In (f), both images align well with the text. The main difference is in the details of the temple facade. 
    Notice that for most images we observe \ours yields images with high quality and visual appeal, as illustrated in \Cref{app:more_images}. 
\end{itemize}
Future work may explore if these challenges can be addressed by following extensions to \ours: (1) \ours could be used in tandem with RL-based method and training-free method to further improve text-to-image faithfulness; (2) prompt engineering methods in DALL-E 3~\cite{BetkerImprovingIG} may help rewriting challenging prompts into simpler ones for models to synthesize; (3) scaling up \ours with a more diverse set of prompts and reward models; (4) mixing \ours with large-scale pre-training on real images.
In summary, as discussed in \S\ref{ssec:limitations}, there is still plenty of headroom to improve.

\clearpage
\section{DSG and Human Rating Evaluation}
\label{dsg-appendix}

\begin{figure}[!ht]
    \centering
    \includegraphics{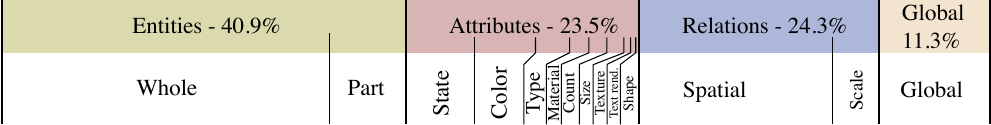}
    \caption{
    {
    Semantic categories contained in DSG.
    \bfit{Entity}: \texttt{whole} (entire entity, e.g., \emph{chair}), \texttt{part} (part of entity, e.g., \emph{back of chair}).
    \bfit{Attribute}: \texttt{color} (e.g., \emph{red book}), \texttt{type} (e.g., \emph{aviator goggles}), \texttt{material} (e.g., \emph{wooden chair}), \texttt{count} (e.g., \emph{5 geese}), \texttt{texture} (e.g., \emph{rough surface}), \texttt{text rendering} (e.g., letters ``Macaroni''), \texttt{shape} (e.g., \emph{triangle block}), \texttt{size} (e.g., \emph{large fence}).
    \bfit{Relation}: \texttt{spatial} (e.g., \emph{A next to B}); \texttt{action} (\emph{A kicks B}).
    \bfit{Global} (e.g., \emph{bright lighting}).
    }}
    \label{fig:dsg-categories}
\end{figure}

\paragraph{Details of DSG-1k benchmark.} \cref{tab:1k_prompt_examples} presents the data sources, quantity, and examples for DSG-1k. \cref{fig:dsg-categories} summarizes the 4 broad and 14 detailed semantic categories covered in the benchmark. 

Like TIFA, DSG~\citep{dsg} falls into the Question Generation / Answering (QG/A) alignment evaluation framework. Unlike TIFA, DSG introduces a linguistically motivated~\cite{donald-davidson-1965,donald-davidson-1967a,donald-davidson-1967b} question generation module to ensure the questions generated to hold 4 reliability traits: a) \emph{atomic}: only queries about 1 semantic detail, for unambiguous interpretation; b) \emph{unique}: no duplicated questions; c) \emph{dependency-aware}: prevent invalid queries to VQA/human answerers, e.g. if the answer to a parent question ``is there a bike?'' is negative, then the child question ``is the bike blue?'' will not be queried; d) \emph{full semantic coverage}: dovetailing the semantic content of a prompt, no more no less. DSG is powered by a large variant of PaLM 2~\citep{anil2023palm} for QG and the SoTA VQA module PaLI~\citep{PaLI} for QA. For our evaluation task, we adopt DSG-1k (DSG's 1,060 benchmark prompt set) which covers a balanced set of diverse semantic categories and writing styles -- including 4 broad categories (e.g. entity/attribute/etc.) and 14 detailed categories (e.g. color/counting/texture/etc.). 

\begin{table}[b]
\newcommand{\parwidth}{9cm}
    \begin{center}
    \resizebox{0.99\textwidth}{!}{
    \begin{tabular}{llcl}
        \toprule
        \textbf{Feature} & \textbf{Source} & \textbf{Sample} & \textbf{Example} \\
        \midrule

        Assorted categories & \makecell[l]{TIFA160~\citep{hu2023tifa}} & 160 & ``A Christmas tree with lights and teddy bear''\\
        
        \midrule
        \multirowcell{2}[-2.5ex][l]{Paragraph-type\\captions} & \makecell[l]{Stanford paragraphs~\citep{krause2016paragraphs}} & 100 & \makecell[{{p{\parwidth}}}]{``There is a cat in the shelf. Under the shelf are two small silver barbels. On the shelf are also DVD players and radio. Beside the shelf is a big bottle of white in a wooden case.''}\\[1.4em]
        
        & \makecell[l]{Localized Narratives~\citep{PontTuset_eccv2020}} & 100 & \makecell[{{p{\parwidth}}}]{``In this picture I can see food items on the plate, which is on the surface. At the top right corner of the image those are looking like fingers of a person.''}\\
        
        \midrule
        Counting & \makecell[l]{CountBench~\citep{paiss2023countclip}} & 100 & \makecell[{{p{\parwidth}}}]{``The view of the nine leftmost moai at Ahu Tongariki on Easter Island''}\\
        
        \midrule
        Relations & \makecell[l]{VRD~\citep{lu2016VRD}} & 100 & \makecell[{{p{\parwidth}}}]{``person at table. person has face. person wear shirt. person wear shirt. chair next to table. shirt on person. person wear glasses. person hold phone''}\\
        
        \midrule
        \multirowcell{2}[-2.5ex][l]{Written by\\T2I \textit{real} users} & \makecell[l]{DiffusionDB~\citep{wangDiffusionDBLargescalePrompt2022}} & 100 & \makecell[{{p{\parwidth}}}]{``a painting of a huangshan, a matte painting by marc simonetti, deviantart, fantasy art, apocalypse landscape, matte painting, apocalypse art''} \\[1.4em]
        
        & \makecell[l]{Midjourney-prompts~\citep{midjourneyprompts2022}} & 100 & \makecell[{{p{\parwidth}}}]{``furry caterpillar, pupa, screaming evil face, demon, fangs, red hands, horror, 3 dimensional, delicate, sharp, lifelike, photorealistic, deformed, wet, shiny, slimy''}\\
        
        \midrule
        Human poses & \makecell[l]{PoseScript~\citep{posescript2022}} & 100 & \makecell[{{p{\parwidth}}}]{``subject is squatting, torso is leaning to the left, left arm is holding up subject, right arm is straight forward, head is leaning left looking forward''}\\
        
        \midrule
        \makecell[l]{Commonsense-\\defying} &  \makecell[l]{Whoops~\citep{bitton2023whoops}} & 100 & \makecell[{{p{\parwidth}}}]{``A man riding a jet ski through the desert''} \\

        \midrule
        Text rendering & \makecell[l]{DrawText-Creative~\citep{Liu2022CharacterAwareMI}} & 100 & \makecell[{{p{\parwidth}}}]{``a painting of a landscape, with a handwritten note that says `this painting was not painted by me'''} \\
         
        \bottomrule
    \end{tabular}
    }
    \caption{{
    \textbf{DSG-1k overview.}
    To comprehensively evaluate T2I models,
    DSG-1k provides 1,060 prompts covering diverse skills and writing styles sampled from different datasets.
    }}
    \label{tab:1k_prompt_examples}
    \end{center}
\end{table}

\begin{figure}[!ht]
    \centering
    \includegraphics[width=0.7\linewidth]{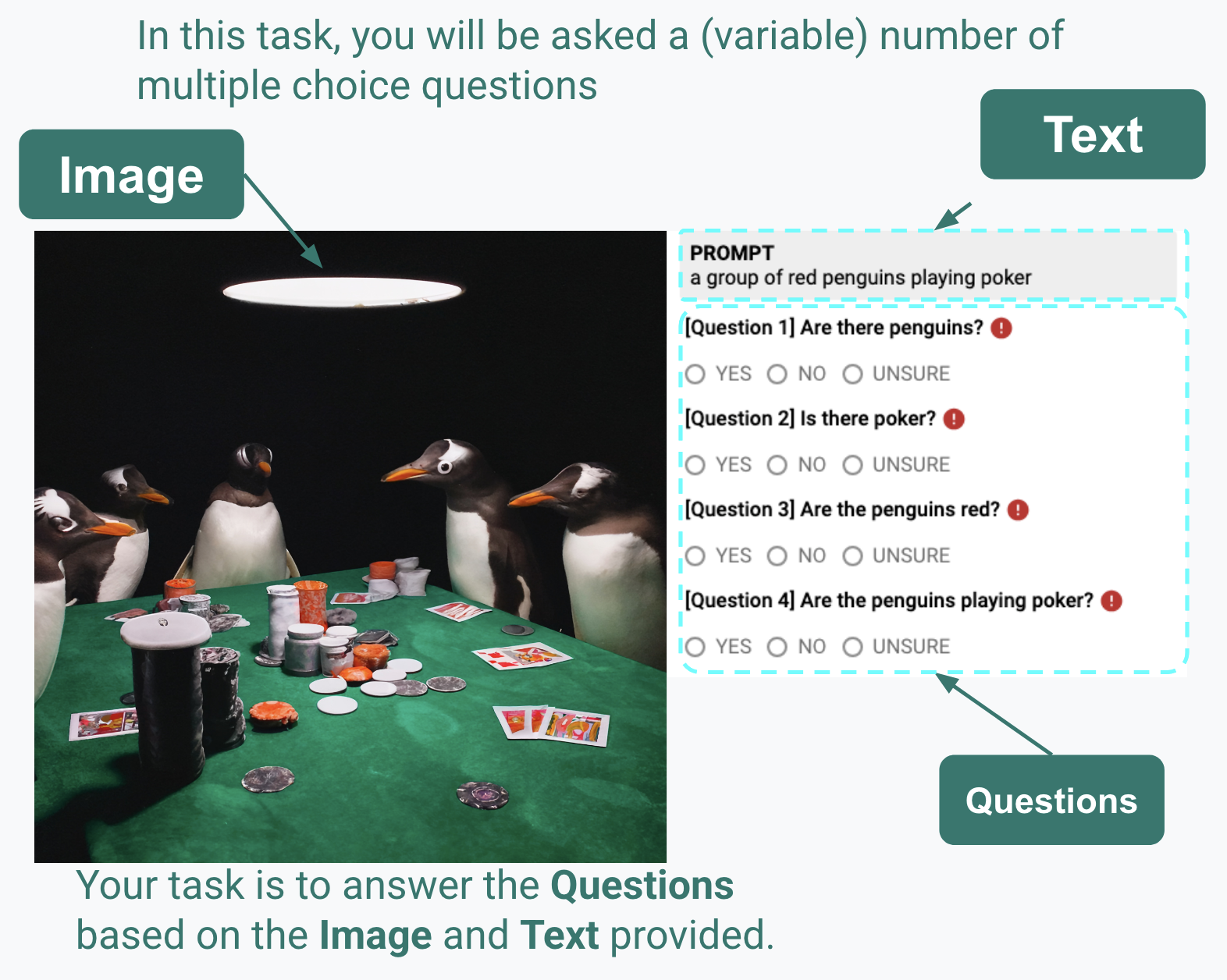}
    \caption{Annotated example of DSG human evaluation query.}
    \label{fig:dsg-human-eval-guide}
\end{figure}

\paragraph{Human QA protocol.} For human evaluation, we elicit 3 rating responses per prompt/question set (with $\sim$8 questions per set on average, and a total of 8183 questions). \cref{fig:dsg-human-eval-guide} exemplifies the UI the human raters see. \cref{fig:dsg-human-eval-protocol} presents the annotation instructions used to guide the raters. The inner-annotator agreement for this study is 0.684.  While the raters respond with \texttt{YES/NO/UNSURE}, we find it to be practically useful to numerically convert the answers -- 1.0 point for \texttt{YES}, 0 for \texttt{NO}, and 0.5 for \texttt{UNSURE} as partial credit, with the justification that if a semantic detail \emph{can potentially} be grounded in an image yet not necessarily so (e.g. ``does this man dress like an engineer?'' image: a male in a plain shirt; ``is this a cat'' image: a blob that \emph{may} be interpreted as a cat), partial credit is fair for not completely failing.

\begin{figure}[h]
\framebox{%
\begin{minipage}[c]{\linewidth}
\footnotesize
\ttfamily

\bfit{INSTRUCTION}

Given an image, a question, and a set of choices, choose the correct choice according to the image content. 

All the questions are formulated as binary: ``YES'' / ``NO''
with an additional option ``UNSURE''

Select ``UNSURE'' if you think the image does not provide
enough information for you to answer the question.
\\

\bfit{NOTES}
\begin{itemize}
    \item Some images may be of low quality. In such cases, please just select the choice according to your intuition. For ambiguous cases, for example, the question is ``is there a man?'', and the image contains a human but it is unclear whether the human is a man, answer ``no''.
    \item If a question assumes something incorrect, select ``UNSURE''.
\end{itemize}

\end{minipage}
}
\caption{Summary of the human annotation instruction for DSG-1k QA.}
\label{fig:dsg-human-eval-protocol}
\end{figure}

\clearpage
\begin{figure}[t]
    \centering
    \includegraphics[width=0.8\linewidth]{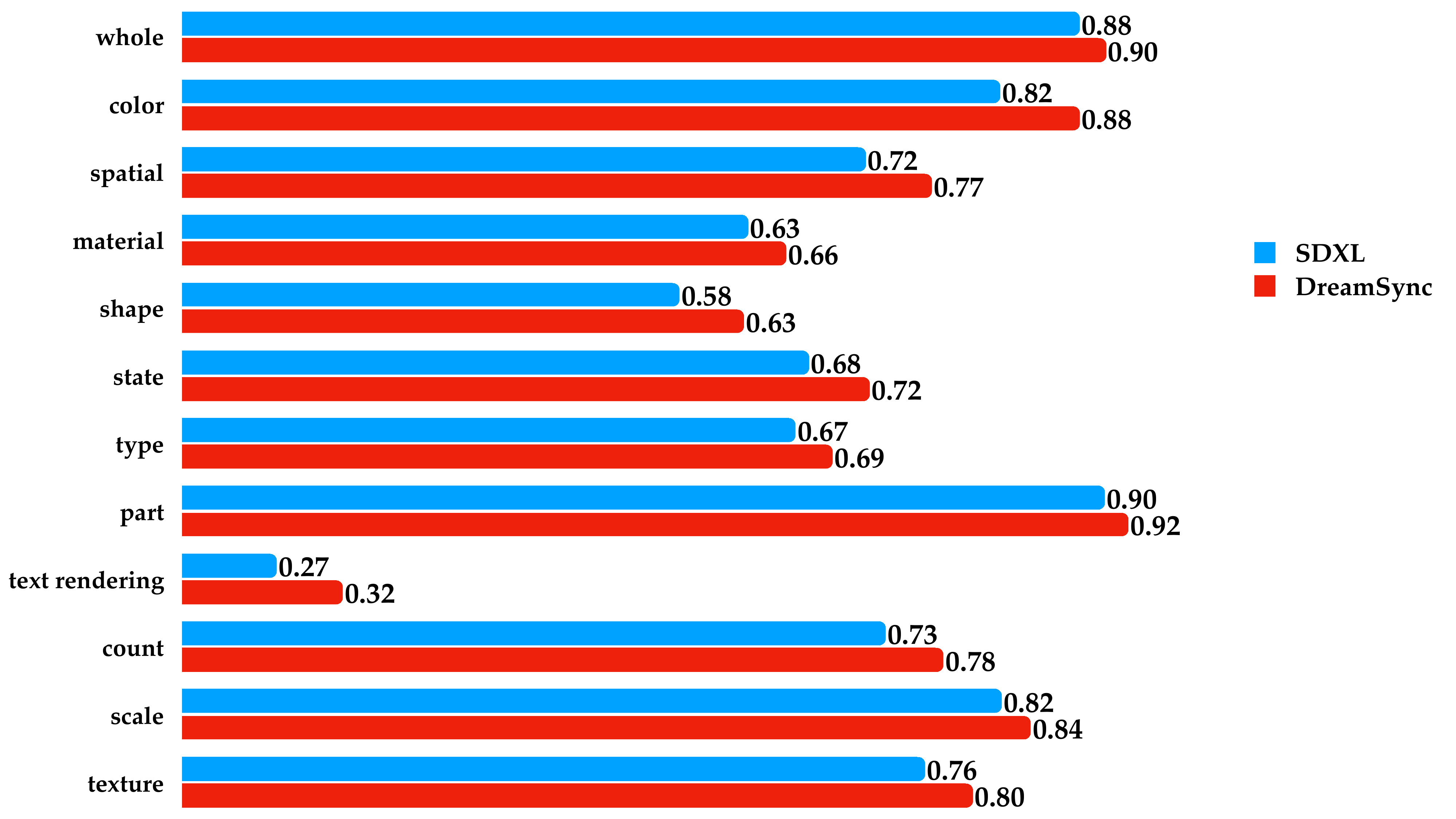}
    \caption{Detailed Human Evaluation Results on DSG-1K. By applying \ours upon SDXL, the human evaluation of alignments improved on all categories. }
    \label{fig:human_eval_detailed}
\end{figure}

\paragraph{Detailed Human Evaluation Results on DSG-1K.} We present detailed evaluations on DSG-1K by semantic categories listed in \Cref{fig:dsg-categories}. The results are shown in \Cref{fig:human_eval_detailed}. By applying \ours upon SDXL, the human evaluation on alignments improved on all categories.

\begin{figure}[b]
    \centering
    \includegraphics[width=0.6\linewidth]{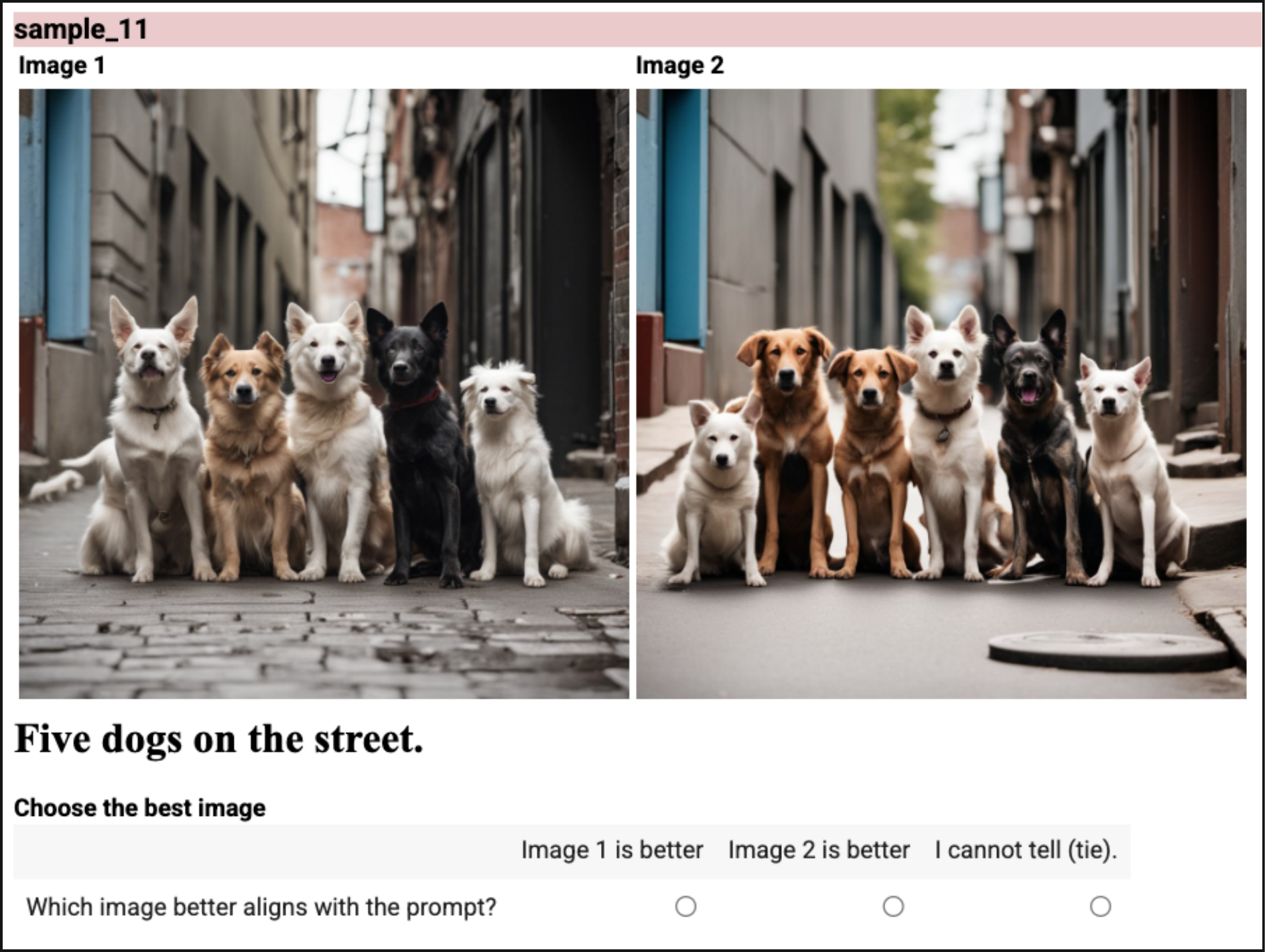}
    \caption{Example human rater screen for the author raters. We display two image side-by-side, where we randomize which is Image 1 vs.~Image 2. We ask a single question to the raters, referencing the prompt that is displayed below the images. The raters were given that instructions \texttt{``Rate images based on how many `components' of the prompt are captured in the image. If both images depict every part of the prompt, then you should choose "I can't tell (tie)." Otherwise, the image with more correct components is better.''} The raters were also shown four examples, with desired ratings and an explanation for the choices.}
    \label{fig:side-by-side-human-eval-appendix}
\end{figure}

\paragraph{Single-Question Human Evaluation.} Besides the large-scale human annotation, we also did a light-weight single-question human evaluation for text prompt alignment. This study was completed by three of the paper's authors. Although this study yields a quite low inter-annotator agreement, we hope it would provide valuable insights on how to set up human evaluation for measuring textual faithfulness of generated images. For this study, we generated one image with SDXL and \ours. See \Cref{fig:side-by-side-human-eval-appendix} for an example rating screen. We randomized the order of the images and prompts. Three authors were asked \texttt{``Which image better aligns with the prompt?''} They could choose Image 1 is better, Image 2 is better, or that they cannot tell (indicating a tie). We use 200 prompts in total with 100 prompts from TIFA and another 100 from DSG.

As mentioned, the inner-annotator agreement was quite low for this study. Only for 42.5\% of the 200 prompts did the human raters all agree in their answers. This is likely due to the fact that it is hard to judge overall prompt alignment directly when given two side-by-side images. Indeed, the majority of prompts led to the raters choosing that they cannot tell which image is better. Using the scoring rules from the DSG study described above (with 1 point going to the model with a direct vote, and with 0.5 going to each model for a tie vote), then we have that \ours scores 50.08 while SDXL scores 49.92.

\paragraph{Key Takeaway from Human Evaluation.} Comparing the fine-grained large-scale human evaluation and the single-question human evaluation, we encourage researchers who are interested in evaluating the text-image alignment to ask annotators detailed and fine-grained questions. It yields significantly better inter-annotator agreement than asking a general single question about alignment. Our large-scale human evaluation with a better agreement suggests that \ours improves the textual faithfulness of SDXL on \texttt{DSG-1k}, resonating with our automatic evaluation.

\clearpage

\section{Randomly-Sampled SDXL+\ours Images} \label{appendix:qualitative_sdxl}
\begin{figure}[h]
    \centering
    \includegraphics[width=0.99\linewidth]{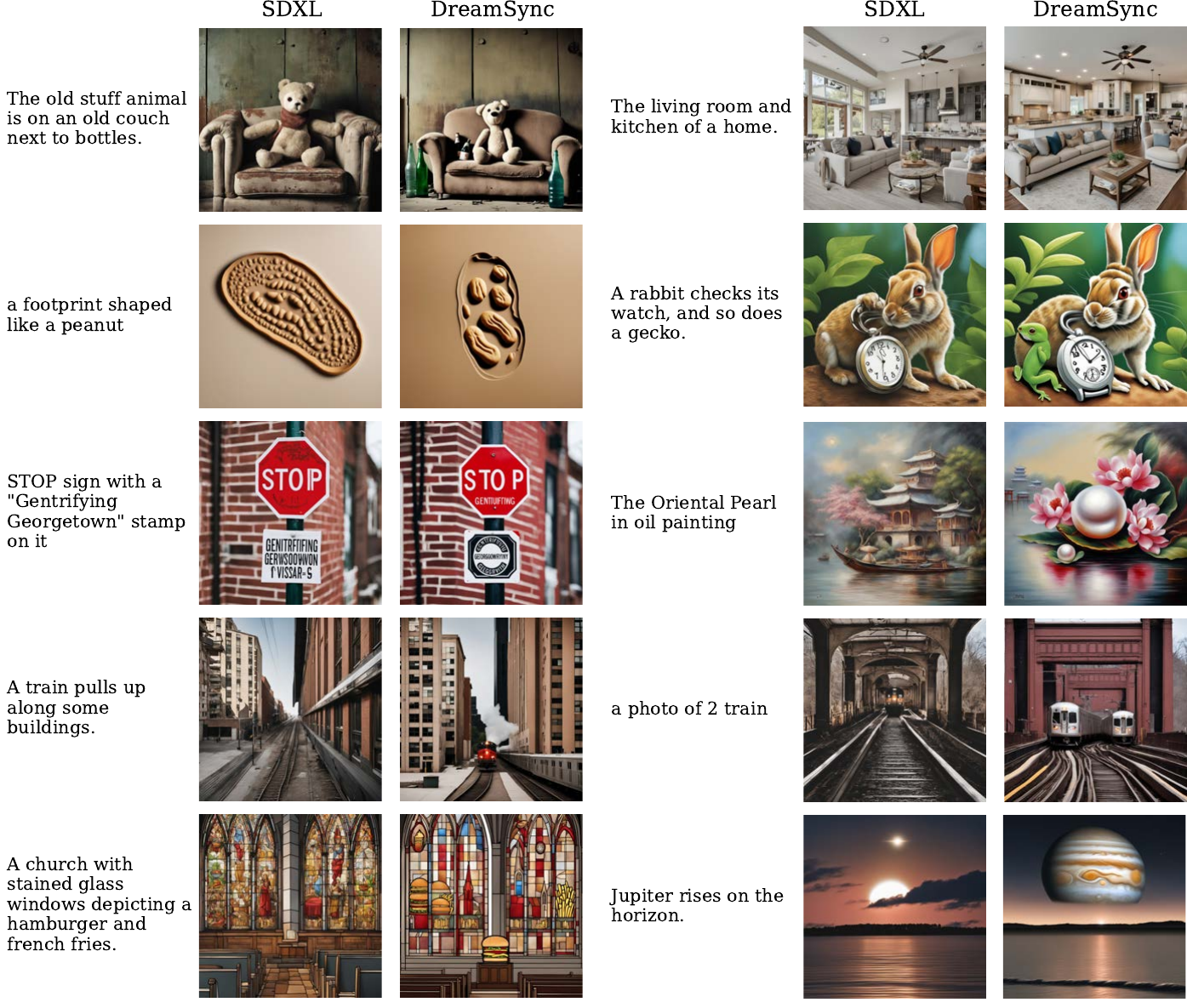}
    \caption{Random samples where \ours are significantly different from SDXL. Both models are sampled with the same seed.}
    \label{fig:sdxl_comparison}
\end{figure}

\begin{figure}[h]
    \centering
    \includegraphics[width=0.99\linewidth]{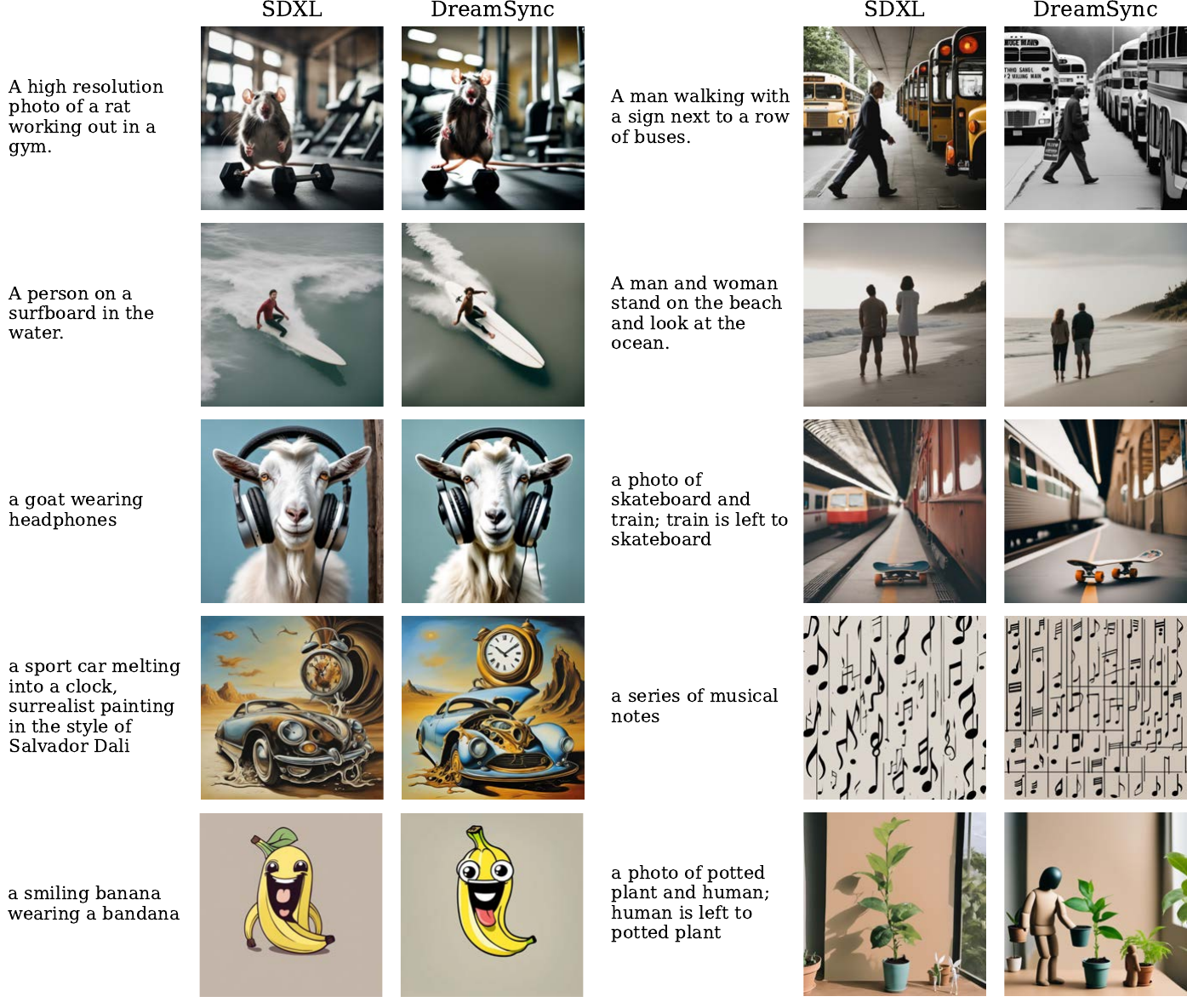}
    \caption{Random samples where \ours barely change SDXL's VQA scores. Both models are sampled with the same seed. We hypothesize that because for simple prompts, SDXL is already good enough to compose them.}
    \label{fig:sdxl_comparison_tie}
\end{figure}

Aside from the failure cases discussed in \Cref{fig:failure_modes_analysis}, 
we would like to showcase more randomly-sampled examples of SDXL and \ours. 
We sample 100 prompts. Among these prompts, \Cref{fig:sdxl_comparison} shows the examples where the VQA scores of applying \ours are significantly different from the base model, SDXL, i.e. the absolute difference of mean score are significantly different: $\big|\mathcal S_M(T, G^{\footnotesize \mathrm{\ours}}(T)) - \mathcal S_M(T, G^{\footnotesize \mathrm{SDXL}}(T))\big| > 0.5$.  Meanwhile, \Cref{fig:sdxl_comparison_tie} presents examples where the \ours does not improve the VQA scores upon SDXL. 

\clearpage
\section{Qualitative Comparison with SD v1.4-based Methods in \Cref{tab:main_result}} \label{appendix:qualitative_sdv1_4}
\begin{figure}[!h]
    \centering
    \includegraphics[width=0.99\linewidth]{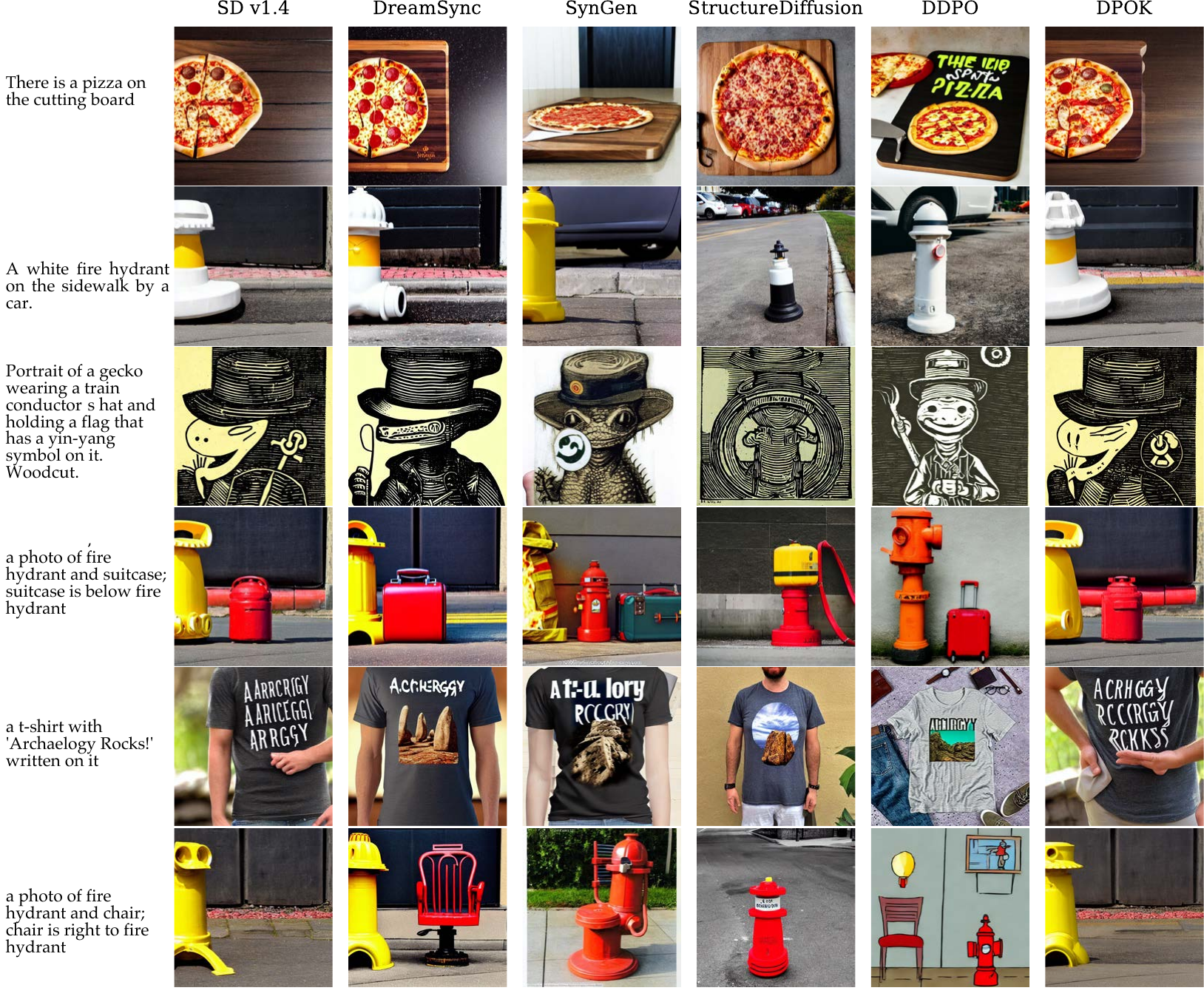}
    \caption{Qualitative Comparison with all models mentioned in \Cref{tab:main_result} with base model SD v1.4. Images are generated with the same seed. \ours improves the base model's alignment to prompts. Unlike RL-based method (e.g. DDPO), \ours does not introduce biases to cartoon. Unlike training-free methods (e.g. SynGen and StructureDiffusion), \ours does not degrade image aesthetics.}
    \label{fig:sdv1_4_comparison}
\end{figure}

Among the 6 examples shown in \Cref{fig:sdv1_4_comparison}, \ours has 3 absolute successes, wheres SynGen, DDPO and StructureDiffusion each has 2, DPOK has 1 and the base model SD v1.4 has 0 absolute success. These results match well with \Cref{tab:main_result}.

\clearpage
\section{Technical Details for Reproducing \ours}
\begin{table}[h]
    \centering
    \begin{tabular}{c|c}
    \toprule
    \multicolumn{2}{c}{Sampling} \\
    \midrule
    Number of Inference Steps & 50 \\
    LoRA $\alpha$ & 0.5 \\
    Prompts per Iteration & 10,000 \\
    Images per Prompt & 8 \\
    Sampling Precision & FP16 \\
    \midrule \midrule
    \multicolumn{2}{c}{Filtering} \\
    \midrule
    $\theta_\mathrm{VQA}$ & 0.9 \\
    $\theta_\mathrm{Aesthetics}$ & 0.6 \\
    Percentage of Prompt-Image Pairs Passing the Filters & 20\% $\sim$ 30\% (see \Cref{fig:chart_improvements}) \\
    Selected Prompt-Image Pairs for Fine-tuning & 2,000 $\sim$ 3,000 \\
    \midrule \midrule
    \multicolumn{2}{c}{LoRA Fine-tuning} \\
    \midrule
        LoRA Rank & 128  \\
        Initial Learning Rate & 0.0001 \\
        Learning Rate Scheduler & Cosine \\
        LR Warmup Steps & 0 \\
        Batch Size & 8 \\
        Gradient Accumulation Steps & 2 \\
        Total Steps & 2,500 \\
        Resolution & 1024 $\times$ 1024 \\
        Random Flip & Yes \\
        Mixed Precision & No (i.e. FP32) \\
        GPUs for Training & 4 $\times$ NVIDIA A6000 \\
        Finetuning Time & $\sim$ 4 Hours \\
    \bottomrule
    \end{tabular}
    \caption{Technical Details for Reproducing \ours with Base Model SDXL.}
    \label{tab:technical_details}
\end{table}

\begin{table}[h]
    \centering
    \begin{tabular}{c|c}
    \toprule
    \multicolumn{2}{c}{Filtering} \\
    \midrule
    $\theta_\mathrm{VQA}$ & 0.85 \\
    $\theta_\mathrm{Aesthetics}$ & 0.5 \\
    \midrule \midrule
    \multicolumn{2}{c}{LoRA Fine-tuning} \\
    \midrule
        Finetuning Time & $\sim$ 1 Hours \\
    \bottomrule
    \end{tabular}
    \caption{Technical Details for Reproducing \ours with Base Model SD v1.4. Same Hyper-parameters as \Cref{tab:technical_details} are omitted.}
    \label{tab:technical_details_sdv14}
\end{table}

\end{document}